\documentclass{article} 
\usepackage{iclr2025_conference,times}

\usepackage{amsmath,amsfonts,bm}

\def\eqref#1{equation~\ref{#1}}

\def\1{\bm{1}}

\DeclareMathAlphabet{\mathsfit}{\encodingdefault}{\sfdefault}{m}{sl}
\SetMathAlphabet{\mathsfit}{bold}{\encodingdefault}{\sfdefault}{bx}{n}

\usepackage{hyperref}
\usepackage{url}

\usepackage[utf8]{inputenc} 
\usepackage[T1]{fontenc}    
\usepackage{hyperref}       
\usepackage{url}            
\usepackage{booktabs}       
\usepackage{amsfonts}       
\usepackage{nicefrac}       
\usepackage{microtype}      
\usepackage{xcolor}         

\usepackage[utf8]{inputenc} 
\usepackage[T1]{fontenc}    
\usepackage{url}            
\usepackage{booktabs}       
\usepackage{amsfonts}       
\usepackage{nicefrac}       
\usepackage{microtype}      
\usepackage{xcolor}         

\usepackage{amsmath}
\usepackage{amssymb}
\usepackage{wrapfig}
\usepackage{subfig}
\usepackage{caption}
\usepackage{color}
\usepackage{algorithm}
\usepackage{algorithmic}
\usepackage{graphicx}

\usepackage{pgfplots}
\usepackage{tikz}
\usepackage{tikz-3dplot}

\usepackage{ragged2e}
\usepackage{fp}

\usepackage{ifthen}

\usepackage{tikz}
\usetikzlibrary{shapes.geometric, arrows, calc,automata,positioning,decorations,decorations.text}

\usetikzlibrary{mindmap,snakes}

\usepackage{multirow}

\usepackage{algorithm}
\usepackage{algorithmic}
\usepackage{graphicx}
\usepackage[english]{babel}
\usepackage{array}
\usepackage[export]{adjustbox}

\usepackage{nicefrac}

\usepackage{xcolor}

\usepackage{colortbl}
\usepackage{makecell}
\usepackage{pifont}

\usepackage{booktabs}

\newcommand{\cmark}{\ding{51}}%
\newcommand{\xmark}{\ding{55}}

\graphicspath{{images/}}

\usepackage[font=small]{caption}

\usepackage{mathptmx}
\usepackage{bm}

\usepackage[outline]{contour}

\definecolor{mycitecolor}{rgb}{0.005, 0.3, 0.7}

\usepackage{url}

\colorlet{accclr}{white!20!black}
\colorlet{flopclr}{green!40!blue}
\colorlet{flopclr}{black!10!flopclr}

\colorlet{baseclr}{gray}
\colorlet{rwclr}{white!90!baseclr}
\colorlet{cellclr}{white!90!baseclr}

  \colorlet{dotaclr}{white}
\colorlet{dotbclr}{magenta}

\newcommand{\bdota}{\textcolor{dotaclr}{$\bullet$}}
\newcommand{\bdotb}{\textcolor{dotbclr}{$\bullet$}}

\colorlet{xmarkclr}{cyan}
\colorlet{cmarkclr}{magenta}

\newcommand{\buac}[1]{\textcolor{#1}{\raisebox{0.22ex}{\protect\contour{#1}{$\uparrow$}}}}  
\newcommand{\bdac}[1]{\textcolor{#1}{\raisebox{0.22ex}{\protect\contour{#1}{$\downarrow$}}}}  

\definecolor{mylinkcolor}{rgb}{0.0, 0.2, 1.0}
 
\definecolor{suppcolor}{rgb}{0.005, 0.3, 0.5}

\FPeval{\tabscale}{0.98}

\newcommand{\OursNet}{CoSNet}
\newcommand{\OursNetFULL}{Columnar Stage Network}
\newcommand{\OursNetHighlight}{\textbf{Co}lumnar \textbf{S}tage \textbf{Net}work}

\title{Designing Concise ConvNets with \\Columnar Stages} 

\author{Ashish Kumar \\
ScoreLabsAI\\
Atlanta, USA \\
\texttt{ashishkumar@gmail.com} \\
\And
Jaesik Park\thanks{Corresponding author} \\
Seoul National University \\
Seoul, South Korea \\
\texttt{jaesik.park@snu.ac.kr} \\
}

\iclrfinalcopy 

\begin{document}

\maketitle

\vspace{-2ex}
\begin{abstract}
\vspace{-0.5ex}
 In the era of vision Transformers, the recent success of VanillaNet shows the huge potential of simple and concise convolutional neural networks (ConvNets). Where such models mainly focus on runtime, it is also crucial to simultaneously focus on other aspects, e.g., FLOPs, parameters, etc, to strengthen their utility further. To this end, we introduce a refreshing ConvNet macro design called \OursNetHighlight{}~(\OursNet{}). \OursNet{} has a systematically developed simple and concise structure, smaller depth, low parameter count, low FLOPs, and attention-less operations, well suited for resource-constrained deployment. The key novelty of \OursNet{} is deploying parallel convolutions with fewer kernels fed by input replication, using columnar stacking of these convolutions, and minimizing the use of $1 \times 1$ convolution layers. Our comprehensive evaluations show that \OursNet{} rivals many renowned ConvNets and Transformer designs under resource-constrained scenarios. Code: \textcolor{mylinkcolor}{\url{https://github.com/ashishkumar822/CoSNet}}.

\end{abstract}
\vspace{-1.5ex}

\section{Introduction}
\label{sec:intro}

In the past decade, there has been enormous study in the neural network architectures \cite{alexnet, vgg}, demonstrating that different information paths \cite{resnet, densenet, googlenet, efficientnet, resnext} can affect the performance. However, as highlighted in recent VanillaNet~\cite{vanillanet}, due to the increased network complexity, the primary source of runtime bottleneck would be the off-chip memory traffic apart from the main computations because GPUs are constantly becoming more powerful.

The issue is prevalent in more advanced models, such as ConvNext~\cite{convnext}, CoatNet~\cite{coatnet}, ViT~\cite{vit}, etc., due to the indirect information paths or the attention mechanism that requires frequent memory reordering. Hence, despite these models being far ahead of their simpler counterparts~\cite{resnet,alexnet}, there are still opportunities to develop concise models for better accuracy, runtime, and resource tradeoffs.


Efforts in this direction are noteworthy. For example, RepVGG~\cite{repvgg} improves runtime via structural parameterization. ParNet \cite{nondeepnetwork} reduces depth by utilizing multiple shallower network modules. Recent VanillaNet~\cite{vanillanet} merges layers during inference while avoiding branches. These works fall in the paradigm of simplifying ConvNet models for resource-constrained scenarios, in contrast to the advanced ConvNets \cite{coatnet,convnext}, or ViT \cite{vit} focusing on state-of-the-art accuracy.

We are inspired by the utility of the former class of works, i.e., simpler and concise models. However, besides focusing on runtime or depth \cite{vanillanet, repvgg, nondeepnetwork}, we also focus on other ConvNet aspects, such as FLOPs, parameters, depth, computational density, etc.
To this end, we propose a concise model by revisiting the fundamentals of prominent ConvNet designs and define the following key sub-objectives:

$1)$ \textit{Reducing depth:} Network depth refers to the number of layers stacked. More depth means more sequential operations, thus more latency and wastage of parallel computing elements (GPU cores).
\\
$2)$ \textit{Controlled parameter growth:} Reducing depth to achieve lower latency leads to an increased number of parameters \cite{vanillanet, nondeepnetwork}, thus necessitating parameter control while having short depth.
\\
$3)$ \textit{Low branching:} Network branching increases memory requirements to hold intermediate tensors and also increases memory access cost to account for the branched operations.
\\
$4)$ \textit{High computational density:} A layer must have a high computing density since fewer computations per layer waste the parallel computing cores, e.g., depthwise convolutions~\cite{mobilenetv1} have less computation density and high memory access cost compared to the dense convolutions~\cite{vgg}.
\\
$5)$ \textit{Uniform primitive operations:} Maintaining a uniform convolution kernel size throughout the network and branches is desirable so that computations can be packed into minimum GPU transactions.

This leads to a concise refreshing ConvNet design (Figure~\ref{fig:netunit}) that shows enhanced performance in various aspects, such as low memory consumption, low memory access costs on parallel computing hardware, smaller depth, minimum branching, lower latency, low parameter count, and reduced FLOPs. The key attributes of \OursNet{}-unit are \textit{parallel columnar convolutions} (Sec.~\ref{sec:pcc}), \textit{input replication} (Sec.~\ref{sec:ir}), and \textit{shallow-deep projections} (Sec.~\ref{sec:proj}), allowing \OursNet{} to perform better than simple ConvNets or rival the advanced designs. The achievements of \OursNet{} emphasize simplicity's importance in effective ConvNet designs.

\section{Related Work}
\label{sec:rel}
%

%
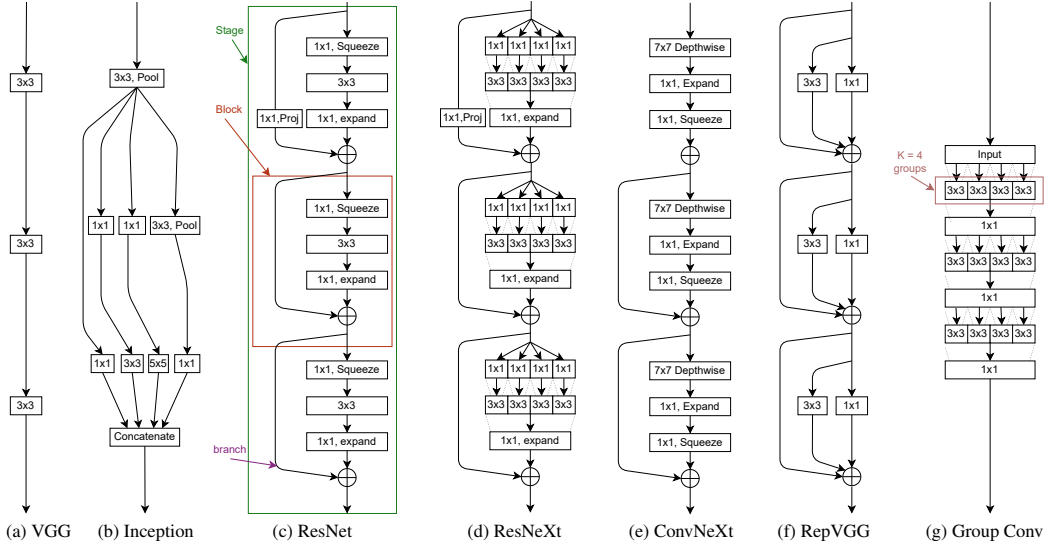
\begin{figure}[t]
\centering

\begin{tikzpicture}

\colorlet{clr}{white!100!gray}
\node (vgg)[draw=clr, rounded corners=0.2ex, line width=0.1ex]{
\includegraphics[scale=0.47]{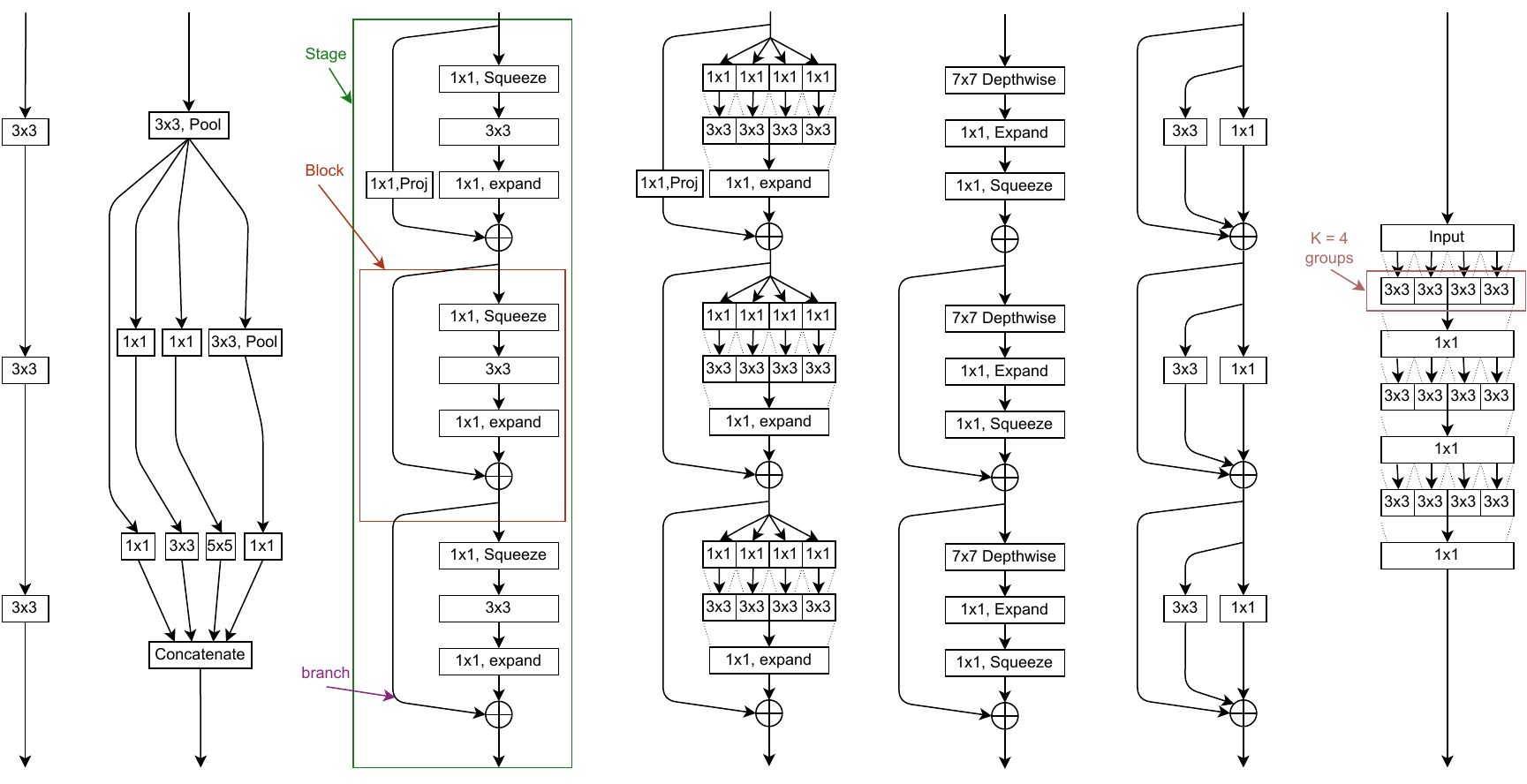}
};

\node (a) [ xshift=-41ex, yshift=-23ex, scale=0.7]{(a) VGG};
\node (b) [xshift=-32ex, yshift=-23ex, scale=0.7]{(b) Inception};
\node (c) [ xshift=-18ex, yshift=-23ex, scale=0.7]{(c) ResNet};
\node (d) [ xshift=-1ex, yshift=-23ex, scale=0.7]{(d) ResNeXt};
\node (e) [ xshift=13ex, yshift=-23ex, scale=0.7]{(e) ConvNeXt};
\node (f) [ xshift=25ex, yshift=-23ex, scale=0.7]{(f) RepVGG};
\node (g) [ xshift=38.5ex, yshift=-23ex, scale=0.7]{(g) Group Conv};

\end{tikzpicture}
\subfloat{\label{fig:stages_vgg}}
\subfloat{\label{fig:stages_inception}}
\subfloat{\label{fig:stages_resnet}}
\subfloat{\label{fig:stages_resnext}}
\subfloat{\label{fig:stages_convnext}}
\subfloat{\label{fig:stages_repvgg}}
\subfloat{\label{fig:stages_groupconv}}
\vspace{-2.5ex}
\caption{Design of various representative architectures in the order of their development in the timeline from (a) to (e). Each graph represents a stage of a network operating at a particular resolution.}
\label{fig:stages}
\vspace{-2ex}
\end{figure}

This section provides an overview of representative network designs (Figure~\ref{fig:stages}). The earlier ConvNets~\citep{alexnet, vgg} stacked dense convolutions with an increasing number of channels and decreasing resolution (Figure~\ref{fig:stages_vgg}). Improved versions \citep{resnet, googlenet, resnext} achieve higher accuracy via manually designed blocks (Figure~\ref{fig:stages_resnet}), while \citep{mobilenetv1, shufflenetv2, mobilenetv2, shufflenetv1}, use depthwise convolutions \citep{depthwise} for saving computations, but they are not memory friendly \citep{repvgg}.

ConvNets have also grown from branchless \citep{alexnet, vgg} (Figure~\ref{fig:stages_vgg}) to single branch \citep{resnet} (Figure~\ref{fig:stages_resnet}) to multi-branch \citep{regnet, inception, efficientnet, nasnet} (Figure~\ref{fig:stages_inception}). These models utilize $1\times1$ convolutions frequently, which rapidly increases network depth \citep{resnet, mobilenetv2, efficientnet, shufflenetv1} (Figure~\ref{fig:stages_resnet}-\ref{fig:stages_convnext}). Although beneficial, both large depth and high branching tend to increase the latency, memory requirements, and Memory Access Cost (MAC) \citep{vanillanet} due to the serialized execution of parallel branches \citep{repvgg, highwaynets, efficientnet}.

Recent RepVGG~\citep{repvgg} proposes structural parameterization (SR) to resolve the branching issue. While ParNet \citep{nondeepnetwork} and VanillaNet \citep{vanillanet} reduce depth to achieve lower latency. Efforts to reduce depth increase the parameter count \citep{vanillanet, nondeepnetwork} to match the accuracy of relatively deeper counterparts \citep{resnet}.

Recent Vision Transformers (ViTs) \citep{vit, efficientvit, swint, deit} have attracted huge research interests. As outlined in \citep{coatnet}, the $O(N^2)$-complex attention in ViTs is a notable issue from a data size and resource-constrained viewpoint. This issue continues to inspire improvements in ConvNets. For instance, RepLKNet \citep{replknet} aims to bridge the gap between ViT and CNNs by employing large kernels.
%

The above designs focus on limited aspects, e.g., \citep{resnet, resnext} on the accuracy, \citep{vanillanet, nondeepnetwork} on runtime and depth. To address this research gap, we draw inspiration from the success of VanillaNet-style networks, and instead of pursuing large-scale models, we focus on our sub-objectives (Sec~\ref{sec:intro}) and revisit the representative ConvNets to push the frontier of simple, concise models.
%
\section{\OursNetFULL{}}
\label{sec:method}
\begin{figure}[t]
\centering
\begin{tikzpicture}

\node (unit)[]{
\includegraphics[scale=0.57]{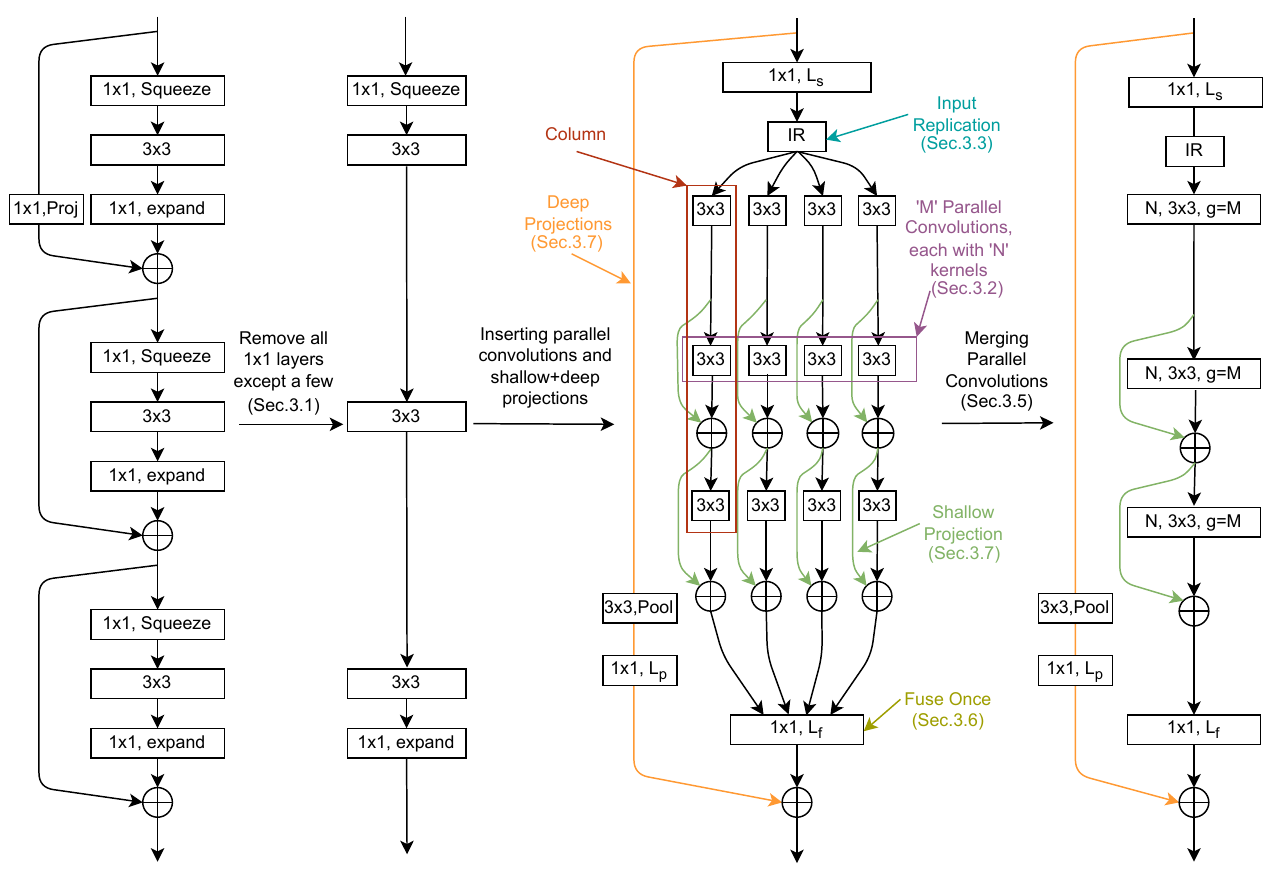}
};
\node (a) [xshift=-32.5ex, yshift=-25.0ex]{\footnotesize (a)};
\node (b) [xshift=-16.8ex, yshift=-25.0ex]{\footnotesize (b)};
\node (c) [xshift=7.0ex, yshift=-25.0ex]{\footnotesize (c)};
\node (d) [xshift=32.0ex, yshift=-25.0ex]{\footnotesize (d)};
\end{tikzpicture}
\subfloat{\label{fig:unit_a}}
\subfloat{\label{fig:unit_b}}
\subfloat{\label{fig:unit}}
\subfloat{\label{fig:optunit}}%
\vspace{-0.5ex}
\caption{Design evolution flow of \OursNet{}-unit. (a) A ResNet \citep{resnet} stage with three blocks. (b) removing all $1\times1$ convolutions except the first of the first block and the last of the last block. (c) detailed design of the \OursNet{}-unit by integrating our design ideas into `(b)', and (d) final optimized \OursNet{}-unit from an implementation viewpoint.}
\label{fig:netunit}
\vspace{-2ex}
\end{figure}
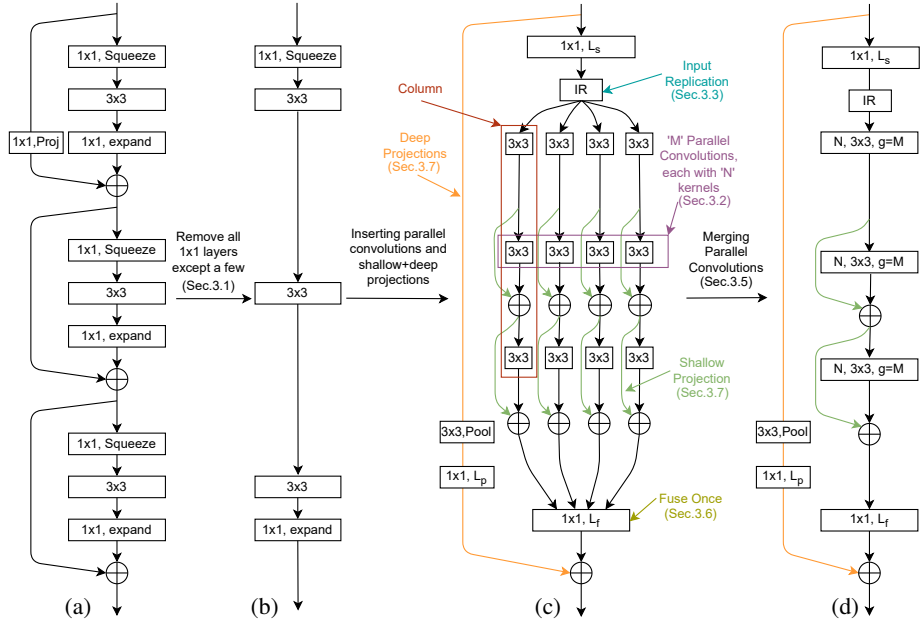

Our approach is a series of improvements motivated by representative ConvNet designs. To understand better, we begin with ResNet~\citep{resnet} as a stepping stone as done in \citep{convnext}. We design the building block of \OursNet{} i.e., \OursNet{}-unit while recalling our sub-objectives: $1)$ reducing depth, $2)$ controlled parameter count, $3)$ high computational density, $4)$ uniform primitive operation, and $5)$ low branching.

\subsection{Avoiding $1\times 1$ for Reducing Depth}

\label{sec:reducing_depth}
The recent works of reducing depth \citep{vanillanet, nondeepnetwork} increase the parameter count to achieve accuracy similar to a deeper network. However, we aim to reduce depth while avoiding a large parameter count, which is a difficult objective. Hence, we handle reducing depth and controlling parameter count separately.

To reduce depth, we identify that $1 \times 1$ convolutions in the ResNet-like designs (Figure~\ref{fig:unit_a}) \citep{resnet, convnext} etc., form almost $66\%$ of depth without improving receptive field due to their pointwise nature \citep{effectivereceptivefield}. Hence, we minimize the number of these layers. Specifically, we use only two $1 \times 1$ convolutions $L_s$ and $L_f$ in a \OursNet{}-unit, where $L_s$ reduces the channel squeezing while $L_f$ performs expansion (Figure~\ref{fig:unit_b}). Then, we stack $l$ number of $3\times3$ convolutions, forming a \textit{column} sandwiched between $L_s$ and $L_f$.

This strategy brings two benefits. \textit{First}, it reduces the overall depth at the same receptive field, e.g., three blocks of ResNet-like design have $9$ layers with three receptive-field governing $3\times3$ layers. In contrast, the proposed design only has $5$ layers, i.e., two $1 \times 1$ and three $3 \times 3$ conv, indicating a notable $45\%$ \emph{depth reduction} with the same receptive field.

\textit{Second}, the reduced depth results in \textit{reduced FLOPs} and \textit{latency} e.g., \OursNet{} performs better than ResNet-$50$ at $50\%$ fewer layers while having relatively fewer parameters, FLOPs, and latency.


\subsection{Parallel Columnar Convolutions for Controlled Parameters.}
\label{sec:pcc}

We propose \textit{Parallel Columnar Convolutions} to handle the large parameter count originating to compensate for the lost non-linearity due to the reduced depth \citep{vanillanet}. In this design, we first deploy $M$ columns in parallel (Figure~\ref{fig:unit}), and crosstalk among columns does not exist, i.e. a convolution of a column can only feed a convolution of the same column. Then, we restrict the number of kernels in a convolution layer of a column to a small number of $N$. This design affects the number of parameters less aggressively when the number of columns increases (see ablations in the supplement). This is a powerful feature of \OursNet{} design, offering controlled growth of parameter count during network scaling. This helps \OursNet{} achieve higher accuracy with fewer parameters.

The idea of the parallel column is based on our hypothesis that multiple kernels with fewer channels can be better than one with large channels. Having $M$ convolutions in parallel with a smaller number of kernels $N$ is equivalent to synthesizing multiple kernels from a large kernel. On the other hand, the idea of smaller $N$ is motivated by the fact that many parallelly operating neurons tend to learn redundant representations while being computationally taxing and causing overfitting. For the same reason, EfficientViT \citep{efficientvit} slices the input channels in its structure. Hence, by keeping $N$ small, we expect to decouple the data patterns learned by the different columns.

In ConvNets, a similar idea was proposed in Inception \citep{googlenet}, then in ResNeXt \citep{resnext}, and then abandoned later as it caused inefficiency. For instance, Inception uses different-sized convolutions and pooling in parallel, which must be executed serially despite being employed in parallel. Also, Inception differs from our columnar architecture since it does not have columns as deep as \OursNet{}.

\subsection{Input Replication}
\label{sec:ir}
In \OursNet{}, all the columns are fed with replicas of the input. We achieve that via a simple \textit{Input Replication} \texttt{IR} operation (Figure~\ref{fig:unit}), which transforms a tensor $\in \mathbb{R}^{C \times H \times W}$ into duplicated one $\in \mathbb{R}^{(M \times C) \times H \times W}$, where $M$ denotes the desired number of the columns. In the \OursNet{}-unit, the \texttt{IR} is applied over the output of the $L_s$ layer to feed each column with the input replica.

Input replication has also been employed in the earlier ResNeXt \citep{resnext}, but notable differences exist. ResNeXt has multiple blocks per stage, and \emph{each block performs} \texttt{IR}, as shown in Figure~\ref{fig:stages_resnext}. Whereas \OursNet{} performs  \texttt{IR} only once. In ResNeXt, \texttt{IR} is performed before $1\times1$ squeeze layer, whereas in \OursNet{}, it is done after the squeeze layer. 

The parallel columnar organization may seem to overlap with widely explored group convolutions \citep{resnext, shufflenetv1}. However, there are two key differences. \textit{First}, group convolution \emph{divides the input channels}, thus defying the objective of \texttt{IR} because now each column receives only a subset of the input channels, thus less information per group, as shown in Figure~\ref{fig:stages_groupconv}. On the contrary, \OursNet{} uses \texttt{IR}, which feeds each column with the replica of the input, thus making the entire input information accessible to each column. This becomes one of the reasons that despite infrequent fusion (Sec.~\ref{sec:fuse_once}), unlike group conv, \OursNet{} still performs better (See ablations in the supplement).

\subsection{Uniform Kernel Size for High Computational Density \& Uniform Primitive Operations.}

The parallel columns of a \OursNet{}-unit can be executed independently; however, this design can be optimized further if all the convolutions in all the columns have uniform kernel size. To this end, we first set the kernel size in all the convolutions to $k \times k$, where $k \in \mathbb{R}_{\geq 3}$. Then, we combine the convolutions of different columns lying at the same level, i.e., the first convolution of each column is combined into one convolution having $M$ batches.

With this optimization, all columns (Figure~\ref{fig:unit}) can be efficiently processed using GPU-based highly optimized Batched-Matrix-Multiply routines, leading to increased computational density, increased GPU utilization, reduced memory access cost \citep{repvgg}, and minimized GPU load-dispatch transactions. Thus resulting in a simplified \OursNet{} design (Figure~\ref{fig:optunit}).
Moreover, since an \OursNet{}-unit is made up mostly of $3\times 3$ convolutions, it well suits the convolution hardware accelerators because they have dedicated support for them, and more chip area can be dedicated to $3\times3$ computational units.

\subsection{Batched Processing for Minimal Branching.} From the previous step, batched processing yields additional benefits, i.e., \OursNet{} becomes uni-branched regardless of training and testing. This reduces memory consumption and access costs, resulting in lower per-iteration training time and increased parallelization. This contrasts with RepVGG \citep{repvgg}, which has a considerable training time. Regarding ASIC development, low branching in \OursNet{} leaves more area on the chip because of the reduced memory requirement to store intermediate tensors. This area can now be dedicated to more computational units.

Although the multi-branch design is beneficial for achieving high accuracy \citep{repvgg} (Figure~\ref{fig:stages_repvgg}), \OursNet{}, despite having minimal branching, effortlessly achieves high accuracy. This is because the core design of \OursNet{}-unit posses multiple branches in the form of columns and short projections (Figure~\ref{fig:unit}). However, due to batched processing \OursNet{}-unit mimics uni-branched behavior. In this way, \OursNet{} takes advantage of both worlds, i.e., eliminated train time complexity due to multiple branches and fast inference during test time without needing structural parameterization \citep{repvgg}. 

\subsection{Fuse Once}
\label{sec:fuse_once}
Finally, the output of all the columns is fused by $L_f$. In ResNeXt (Figure~\ref{fig:stages_resnext}), the output of $3\times3$ convs are fused immediately via a $1\times1$ conv, whereas in \OursNet{}, it is done much later. 
Our fuse once strategy is different from group \citep{shufflenetv1} or depthwise convolutions \citep{mobilenetv1} that are followed by $1\times 1$ (Figure~\ref{fig:stages_groupconv}) to avoid loss of accuracy because each group/channel has too few connections which restrict its learning ability without frequent fusion (\citep{shufflenetv1}, Figure~\ref{fig:stages_groupconv}). This increases network depth and, hence, latency. On the contrary, \OursNet{} is free from this constraint because we increase $N$ as we go deep in \OursNet{} unlike \citep{shufflenetv1}. Hence, each neuron in $M$ columns has a sufficiently large number of connections that enable learning without frequent fusion. We performed an ablation (see supplement) by applying the same strategy as Figure~\ref{fig:stages_groupconv} in \OursNet{}. We observed increased network depth, latency, and decreased accuracy.

\begin{wrapfigure}[11]{r}{0.30\textwidth}
\centering
\vspace{-4ex}
\includegraphics[scale=0.5]{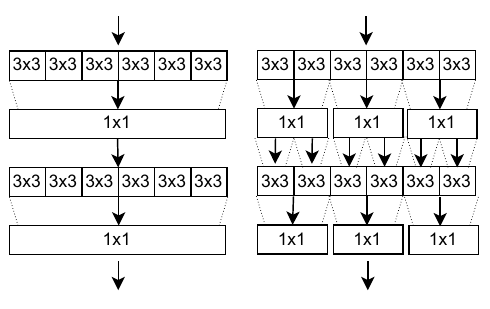}
\vspace{-4.0ex}
\caption{Illustration of Vanilla Frequent Fusion (left) (\citep{shufflenetv1}, Figure~\ref{fig:stages_groupconv}) and Pairwise Frequent Fusion (right).}
\label{fig:pff}
\vspace{-2.5ex}
\end{wrapfigure}

\textit{Pairwise Frequent Fusion (PFF):} Although we aim to reduce $1\times1$ layers as they have a high concentration of most of the network parameters and FLOPs (Sec.~\ref{sec:reducing_depth}), we propose a frequent fusion scheme via $1\times 1$ while avoiding the parameter and FLOPs concentration issue. In this scheme, instead of fusing all the columns simultaneously, we fuse columns only pairwise via $1 \times1$ (Figure~\ref{fig:pff}). This strategy essentially offers several benefits. Firstly, with pairwise fusion, $1 \times 1$ kernel incorporates only a few computations per layer due to small kernel size (fewer channels) while improving network accuracy. Secondly, the latency incorporated due to these layers does not increase the overall latency because of the few computations per layer, hence offers better accuracy with negligible latency overhead ($1-2$ms). See Table~\ref{tab:imagenet_300}. We denote all such \OursNet{} variants as \OursNet{}-PFF.
\subsection{Projections}
\label{sec:proj}
To facilitate better gradient flow during network training, we employ projections introduced by ResNet \citep{resnet} but slightly differently in two ways:

$1)$ \textit{Shallow Range.} These projections are formed between any two layers of a column and promote better gradient flow through the stack of $l$ layers (Figure~\ref{fig:unit}). Since such projections connect only two layers, unlike a stack of layers in ResNet-like designs, these are named shallow ranges.

$2)$ \textit{Deep Range.} These projections are formed between the input and the output of a \OursNet{}-unit. Specifically, the input to \OursNet{}-unit is projected to its output via a $3 \times 3$ pooling layer followed by a $1 \times 1$ convolution $L_p$ whose output is fused with the output of $L_f$  (Figure~\ref{fig:unit}). The pooling operation gathers spatial context by enlarging the receptive field, which is otherwise impossible for $L_p$ alone due to its point-wise nature. 
We call it deep projection because it bypasses the entire columnar structure while combining information from the previous network stages, i.e., multi-layer information fusion, and providing a short alternative path for gradient flow. 

The above projection design helps achieve \OursNet{} better accuracy (see ablations) and is slightly different from the existing ones. First, projection in ResNet-like models \citep{resnet, resnext} is used only in the first block of a stage (shallower), and projection between stages does not exist. Second, projection in these models operates at a stride of $2$. On the contrary, in \OursNet{}, the projection connects two stages (deeper) while operating at unit stride and utilizing pooling to increase the receptive field. 
\subsection{\OursNet{} Instantiation}
\label{sec:designspace}
%

\begin{wrapfigure}[15]{r}{0.32\textwidth}
\centering
\vspace{-4ex}

\begin{tikzpicture}

\node (unit)[]{
\includegraphics[scale=0.45]{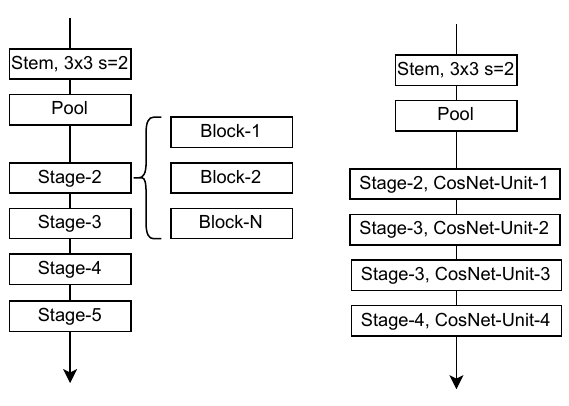}
};
\node (a) [xshift=-6ex, yshift=-11ex]{\scriptsize (a) Existing Models};
\node (b) [xshift=9ex, yshift=-11ex]{\scriptsize (b) \OursNet{}};

\end{tikzpicture}
\subfloat{\label{fig:old_arch}}
\subfloat{\label{fig:arnet_arch}}
\vspace{-3.0ex}
\caption{Macro design of (a) existing networks e.g. \cite{repvgg, resnet, convnext, resnext}, and (b) \OursNet{}. \OursNet{} does not have blocks in its stages.}
\label{fig:macroarch}
\vspace{-2.5ex}
\end{wrapfigure}

A \OursNet{} variant can be instantiated by stacking \OursNet{}-units (Figure~\ref{fig:macroarch}). \OursNet{} does not have the notion of blocks but only has stages in the form of \OursNet{}-unit. This contrasts with existing ConvNets, which have stages, and each stage comprises multiple blocks \citep{nondeepnetwork, resnet, convnext, resnext} e.g., ResNet-$50$ has four stages, having $3$, $4$, $6$, and $3$ blocks respectively (Figure~\ref{fig:stages_resnet}-~\ref{fig:stages_convnext}).

To instantiate a \OursNet{} variant, we follow the tradition of five stages \citep{resnet, vgg}, among which the first (stem) is a $3 \times 3$ convolution with a stride of $2$, while the remaining are the \OursNet{}-units. 

Following ResNet~\citep{resnet}, we set channels of $L_s$ to $64$, which gets doubled at each stage, while the channels of $L_p$ and $L_f$ always equal to $\zeta$ times the channels of $L_s$. We set $\zeta=4$, following \citep{resnet}. To further simplify the instantiation, we set the depth of a column, i.e., $l$ in $k^{th}$ \OursNet{}-unit equal to the number of blocks in the $k^{th}$ stage of a widely used model ResNet-$50$~\citep{resnet}. Summarily, \OursNet{}-unit has \emph{only three hyperparameters: $M, N, l$} which control \OursNet{}'s parameters, depth, latency, and accuracy. Hence, different \OursNet{} variants can be constructed by changing them. Please refer to the supplement for \OursNet{} instance names and ablations on $M, N, l$.

%
\vspace{-1ex}
\section{Experiments}
\label{sec:exp}

We evaluate \OursNet{} on ImageNet \citep{imagenet} dataset consisting of $1.28$M train and $50$k validation images of $1000$ categories. Our training methodology is consistent with recent VanillaNet~\citep{vanillanet}. We use data augmentation techniques in \citep{vanillanet, convnext}. See the appendix at the end of this paper for more details.

\begin{table*}[!t]
\centering

\caption{Evaluation of \OursNet{} on ImageNet \cite{imagenet}. Latency is measured with batch size $1$. `SR' denotes structural parameterization. `PFF' stands for pairwise frequent fusion. See Sec~\ref{sec:fuse_once} for details.}
\label{tab:imagenet_300}
\vspace{-1ex}
\arrayrulecolor{white!60!black}
\setlength{\tabcolsep}{3.0pt}

\scriptsize

\resizebox{0.995\linewidth}{!}
{

\begin{tabular}{l | c | c r r r r c c}

\toprule

Architecture & Type & \#Depth~\bdac{accclr} & \#Params~\bdac{accclr} & FLOPs~\bdac{accclr} & Latency~\bdac{accclr} & FPS~\buac{accclr} & Top-$1$ ($\%$)~\buac{accclr} \\

\midrule

\bdota{}~ResNet-$18$ \cite{resnet} & ConvNet & $18$ &  $11.6$M & $1.83$B & $4$ms & $250$ & $71.1$ \\
\bdota{}~ResNet-$34$ \cite{resnet} & ConvNet & $34$ &  $21.7$M & $3.68$B  & $8$ms & $125$ & $74.1$ \\
\bdota{}~ResNet-$50$ \cite{resnet} & ConvNet & $50$ &  $25.5$M & $4.12$B & $11$ms & $90$ & $76.3$ \\
\bdota{}~ResNet-$101$ \cite{resnet} & ConvNet & $101$ &  $44.5$M & $7.85$B & $15$ms & $67$ & $77.2$ \\
\bdota{}~ResNet-$152$ \cite{resnet} & ConvNet & $152$ &  $60.1$M & $11.50$B & $15$ms & $67$ & $77.8$ \\

\midrule

\bdota{}~ResNeXt-$50$ \cite{resnext} & ConvNet & $50$ &  $25.1$M  & $4.40$B & $11$ms & $90$ & $77.4$ \\
\bdota{}~ResNeXt-$101$ \cite{resnext} & ConvNet & $101$ &  $44.1$M & $8.10$B & $14$ms & $71$ & $78.4$ \\

\midrule

\bdota{}~EfficientNet-B$0$ \cite{efficientnet} & ConvNet & $49$ &  $5.3$M & $0.40$B & $8$ms & $125$ & $75.1$ \\

\midrule 

\bdota{}~RegNetX-$12$GF \cite{regnet} & ConvNet & $57$ & $46.0$M & $12.10$B & $13$ms & $77$ & $80.5$ \\

\midrule

\bdota{}~RepVGG-A$0$ \cite{repvgg} & ConvNet & $22$ & $8.3$M  & $1.46$B & $4$ms & $250$ & $72.4$ \\
\bdota{}~RepVGG-A$0$ \cite{repvgg} w/o SR & ConvNet & $22$ & $9.1$M & $1.51$B & $8$ms & $125$ & $72.4$ \\

\bdota{}~RepVGG-A$1$ \cite{repvgg} & ConvNet & $22$ & $12.7$M  & $2.36$B & $5$ms & $200$  & $74.4$ \\ 
\bdota{}~RepVGG-A$1$ \cite{repvgg} w/o SR & ConvNet & $22$ & $14.0$M & $2.63$B & $7$ms & $143$ & $74.4$ \\ 

\bdota{}~RepVGG-B$0$ \cite{repvgg} & ConvNet & $28$ & $14.3$M & $3.40$B & $5$ms & $200$  & $75.1$ \\ 
\bdota{}~RepVGG-B$0$ \cite{repvgg} w/o SR & ConvNet & $28$ & $15.8$M & $3.06$B & $7$ms & $143$ & $75.1$ \\ 

\bdota{}~RepVGG-A$2$ \cite{repvgg} & ConvNet & $22$ & $25.5$M & $5.12$B & $7$ms & $143$ & $76.4$ \\
\bdota{}~RepVGG-A$2$ \cite{repvgg} w/o SR & ConvNet & $22$ & $28.1$M & $5.69$B & $9$ms & $111$ & $76.4$ \\

\bdota{}~RepVGG-B$3$ \cite{repvgg} & ConvNet & $28$ & $110.9$M & $26.20$B & $17$ms & $58$ & $80.5$ \\
\bdota{}~RepVGG-B$3$ \cite{repvgg} w/o SR & ConvNet & $28$ & $123.0$M & $29.10$B & $22$ms & $45$ & $80.5$ \\

\midrule

\bdota{}~ParNet-L \cite{nondeepnetwork} & ConvNet & $12$ &  $55.0$M & $26.70$B & $23$ms & $43$ & $77.7$ \\
\bdota{}~ParNet-XL \cite{nondeepnetwork} & ConvNet & $12$ &  $85.0$M & $41.50$B & $25$ms & $40$ & $78.5$ \\ 

\midrule

\bdota{}~DeiT-S \cite{deit} & Transformer & $48$ &  $22.0$M & $4.60$B & $15$ms & $66$ & $79.8$ \\
\bdota{}~Swin-T \cite{swint} & Transformer & $96$ &  $28.0$M & $4.50$B & $20$ms & $50$ & $81.1$ \\
\bdota{}~ViTAE-S \cite{vitae} & Transformer & $116$ & $23.6$M & $5.60$B & $24$ms & $41$ & $82.0$ \\

\midrule
\bdota{}~CoAtNet-0 \cite{coatnet} & Hybrid & $64$ & $25.0$M & $4.20$B & $15$ms & $66$ & $81.6$ \\

\midrule

\bdota{}~ConvNeXt-T \cite{convnext} & ConvNet & $59$ & $29.0$M & $4.50$B & $13$ms & $77$ & $81.8$ \\
\bdota{}~ConvNextV$2$-P \cite{convnextv2} & ConvNet & $41$ & $9.1$M & $1.37$B & $11$ms & $90$ & $79.7$ \\
\bdota{}~ConvNextV$2$-N \cite{convnextv2} & ConvNet & $47$ & $15.6$M & $2.45$B & $13$ms & $77$ & $81.2$ \\
\bdota{}~ConvNextV$2$-T \cite{convnextv2} & ConvNet & $59$ & $28.6$M & $4.47$B & $16$ms & $62$ & $82.5$ \\

\midrule

\bdota{}~EfficientViT-M$4$ \cite{efficientvit} & Transformer & $42$ &  $8.8$M & $0.30$B & $6$ms & $166$ & $74.3$ \\
\bdota{}~EfficientViT-M$5$ \cite{efficientvit} & Transformer & $70$ &  $12.4$M & $0.60$B & $7$ms & $142$ & $76.8$ \\

\midrule

\bdota{}~VanillaNet-6 \cite{vanillanet} & ConvNet & $6$  & $32.0$M & $6.00$B  & $6$ms & $167$ & $76.3$ \\
\bdota{}~VanillaNet-8 \cite{vanillanet} & ConvNet & $8$ & $37.1$M & $7.70$B  & $6$ms & $167$ & $79.1$ \\
\bdota{}~VanillaNet-9 \cite{vanillanet} & ConvNet & $9$  & $41.4$M & $8.60$B  & $6$ms & $167$ & $79.8$ \\
\bdota{}~VanillaNet-10 \cite{vanillanet} & ConvNet & $10$ & $45.7$M & $9.40$B  & $7$ms & $142$ & $80.5$ \\

\midrule

\bdota{}~InceptionNeXt-S \citep{inceptionnext} & ConvNet & $48$ &   $49.0$M & $8.40$B  & $18$ms & $55$ & $83.5$ \\

\bdota{}~UniRepLKNet-S \citep{unireplknet} & ConvNet  & $180$ &  $56.0$M & $9.10$B  & $23$ms &  $43$ & $83.9$ \\

\midrule
\bdotb{}~\textbf{\OursNet{}-A$0$} & ConvNet & $26$ &  $8.8$M & $1.25$B & $6$ms & $167$ & $77.1$ \\ 
\bdotb{}~\textbf{\OursNet{}-A$1$} & ConvNet & $26$ & $12.1$M & $1.70$B  & $6$ms & $167$ & $78.2$ \\ 
\bdotb{}~\textbf{\OursNet{}-B$0$} & ConvNet & $26$ & $19.8$M & $3.05$B & $7$ms & $143$ & $79.5$ \\ 
\bdotb{}~\textbf{\OursNet{}-B$1$} & ConvNet & $26$ & $22.0$M & $3.50$B  & $7$ms & $167$ & $79.9$ \\ 
\bdotb{}~\textbf{\OursNet{}-B$2$} & ConvNet & $26$ & $30.0$M & $5.10$B  & $9$ms & $111$ & $81.3$ \\ 
\bdotb{}~\textbf{\OursNet{}-C$1$} & ConvNet & $28$ & $24.4$M & $4.12$B & $7$ms & $143$ & $80.0$ \\ 
\bdotb{}~\textbf{\OursNet{}-C$2$} & ConvNet & $26$ & $38.9$M & $7.09$B & $11$ms & $90$ & $82.1$ \\ 

\midrule
\bdotb{}~\textbf{\OursNet{}-A$1$-PFF} & ConvNet & $38$ & $12.7$M & $1.93$B  & $7$ms & $143$ & $79.7$ \\ 
\bdotb{}~\textbf{\OursNet{}-B$0$-PFF} & ConvNet & $38$ & $21.8$M & $3.44$B & $8$ms & $125$ & $80.6$ \\ 
\bdotb{}~\textbf{\OursNet{}-B$1$-PFF} & ConvNet & $38$ & $25.6$M & $4.08$B  & $8$ms & $125$ & $81.4$ \\ 
\bdotb{}~\textbf{\OursNet{}-B$2$-PFF} & ConvNet & $38$ & $34.3$M & $5.91$B  & $10$ms & $100$ & $82.7$ \\ 
\bdotb{}~\textbf{\OursNet{}-C$1$-PFF} & ConvNet & $42$ & $27.3$M & $4.75$B & $8$ms & $125$ & $81.3$ \\ 
\bdotb{}~\textbf{\OursNet{}-C$2$-PFF} & ConvNet & $38$ & $44.5$M & $8.27$B & $13$ms & $77$ & $83.7$ \\

\bottomrule

\end{tabular}
}
\vspace{-5ex}
\end{table*}

\subsection{Advanced ConvNets and Vision Transformers}
\label{sec:big_cnn}
%

\begin{figure}[t]

\centering

\begin{tikzpicture}

\colorlet{clr}{white!100!gray}
\node (1)[draw=none, xshift=0ex, yshift=0ex]{
\includegraphics[scale=0.68]{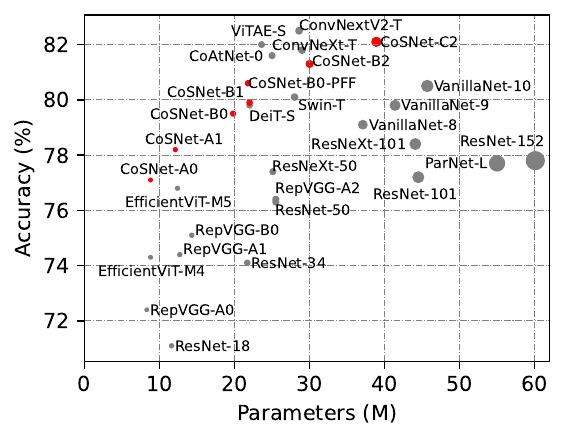}
};
\node (2)[draw=none, xshift=45ex, yshift=0ex]{
\includegraphics[scale=0.68]{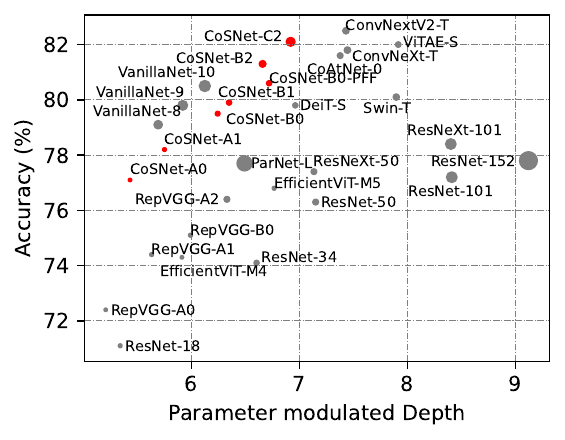}
};
\node (3)[draw=none, xshift=0.5ex, yshift=-32ex]{
\includegraphics[scale=0.68]{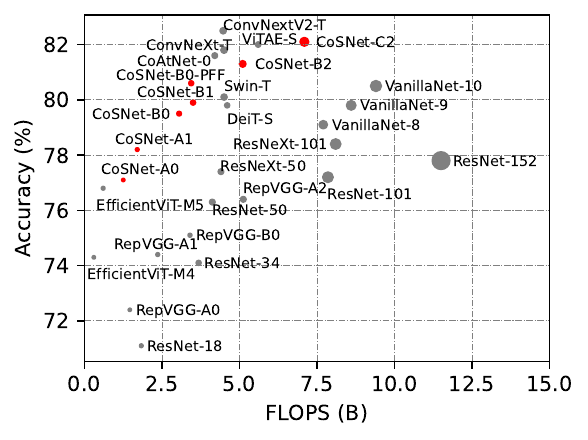}
};
\node (4)[draw=none, xshift=45ex, yshift=-32ex]{
\includegraphics[scale=0.68]{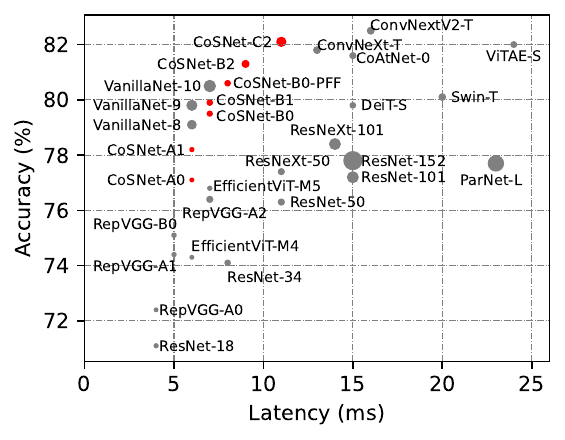}
};
\node (a) [below of=1, xshift=-18ex, yshift=-7.5ex, scale=0.9]{(a)};
\node (b) [below of=2, xshift=-18ex, yshift=-7.5ex, scale=0.9]{(b)};
\node (c) [below of=3, xshift=-18ex, yshift=-7.5ex, scale=0.9]{(c)};
\node (d) [below of=4, xshift=-18ex, yshift=-7.5ex, scale=0.9]{(d)};

\end{tikzpicture}
%
%
%
\vspace{-3.5ex}
%
\def\circ{\protect\raisebox{0.2ex}{\protect\tikz{\protect\node[draw=red, fill=red,scale=0.5, circle]{};}}}
\def\circg{\protect\raisebox{0.2ex}{\protect\tikz{\protect\node[draw=gray, fill=gray,scale=0.5, circle]{};}}}
\caption{Comparing the proposed \OursNet{} with representative models. Models in `\protect\circ{} and `\protect\circg{} refers to \OursNet{} and existing models respectively. \OursNet{} has lower parameters, lower FLOPs, while depth of \OursNet{} is not unnecessarily large. The size of the circle is proportional to the parameter count.}

\label{fig:mde}
\vspace{-1ex}

\end{figure}
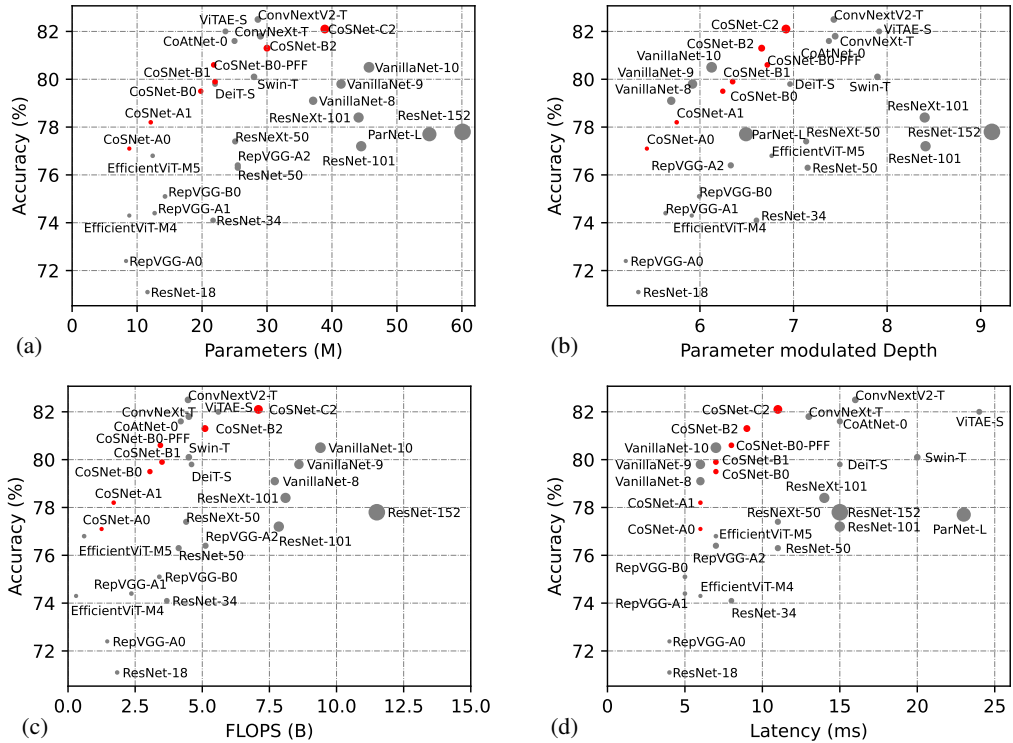

\noindent\textbf{\OursNet{} \textit{vs} recent EfficientViT \citep{efficientvit}}
As shown in Table~\ref{tab:imagenet_300} and Figure~\ref{fig:mde}, \OursNet{} is less deep and runs $60\%$ faster than EfficientViT Transformer while exhibiting better accuracy, e.g., EfficientVit-M$4$ vs \OursNet{}-A$0$. EfficientVit is another example of lower FLOPs that do not guarantee lower latency. Even the \OursNet{}-A$1$-PFF variant is still relatively shallower than EfficientVit while delivering better accuracy.

\noindent\textbf{\OursNet{} \textit{vs} DeiT \citep{deit}}
From Table~\ref{tab:imagenet_300}, \OursNet{}-B$1$ is almost $50\%$ less deep, has $23\%$ fewer params, and runs $60\%$ faster than DeiT Transformer while exhibiting slightly better accuracy. With PFF, \OursNet{}-B$0$-PFF performs better in terms of accuracy, depth, and runtime.

\noindent\textbf{\OursNet{} \textit{vs} advanced mid-range ConvNets and Transformers}
\OursNet{}-B$2$ is $72\%$ less deeper, $55\%$ faster, and $1.2\%$ more accurate than the popular Swin Transformer~\citep{swint}. It is also $55\%$ less deeper, $30\%$ faster with slightly lower accuracy than the popular ConvNeXt~\citep{convnext}. Moreover, \OursNet{}-C$2$ rivals the latest ConvNext-v$2$-T \citep{convnextv2} with similar accuracy but higher speed and smaller depth.

\OursNet{}-B$2$, C$1$, C$2$ models rivals advanced Transformers, such as ViTAE-S~\citep{vitae} and hybrid models, such as CoAtNet-0~\citep{coatnet}. With similar parameter counts and accuracy, our models show faster inference speed. The competitive tradeoffs offered by \OursNet{} show the significance of concise models.

\subsection{Comparison with Standard ConvNets} We show that \OursNet{} achieves efficiency in multiple aspects in a large spectrum of models while being simpler during training and inference and offering competitive trade-offs relative to the rival network. See Table~\ref{tab:imagenet_300} for the comparison. Figure~\ref{fig:mde} plots the trends regarding various aspects.

\noindent\textbf{\OursNet{} \textit{vs} recent VanillaNet~\citep{vanillanet}}. 
\OursNet{} rivals recent ConvNet design, VanillaNet. VanillaNet is shallow and mainly focuses on latency. Our \OursNet{}-A$0$ shows similar latency at fewer parameters, fewer FLOPs, and high accuracy compared with VanillaNet-$6$ (Table~\ref{tab:imagenet_300}). 

\noindent\textbf{\OursNet{} \textit{vs} recent ParNet \citep{nondeepnetwork}.} 
\OursNet{} outperforms recent non-deep ParNet that focuses on lower latency (Table~\ref{tab:imagenet_300} R4). \OursNet{} is uni-branched, while ParNet has multiple shallow branches which serialize the computations, thus making them deeper virtually.

\noindent\textbf{\OursNet{} \textit{vs} RepVGG \citep{repvgg}.}
RepVGG offers a plain VGG-like \citep{vgg} structure via Structural Reparameterization (SR). However, its training complexity is high due to a large number of parameters and three branches at each layer (Figure~\ref{fig:stages_repvgg}). Hence, we show its performance with and without SR.

Compared with the RepVGG family, \OursNet{} offers considerably lower complexity during training and testing, thanks to its parallel columnar convolutions. In addition, \OursNet{} has fewer parameters and fewer FLOPs while offering similar speeds with higher accuracy. For instance, \OursNet{}-B$2$ is better than RepVGG-B$3$ at similar depth, $73\%$ fewer parameters, $80\%$ lesser FLOPs while running faster. This shows the significance of parallel columns of \OursNet{} that during model scaling, parameter count does not grow rapidly.

\noindent\textbf{\OursNet{} \textit{vs} EfficientNet \citep{efficientnet}.} 
Although we do not aim for a mobile regime in this paper, we show that having fewer parameters and FLOPs does not guarantee faster speeds. As shown in Table~\ref{tab:imagenet_300}, EfficientNet-B$0$ has $50\%$ fewer parameters and $77\%$ fewer FLOPs, but is $50\%$ deeper, and runs $37\%$ slower. By exploring the design space, \OursNet{} can be extended to the mobile regime.

\noindent\textbf{\OursNet{} \textit{vs} ResNet \citep{resnet} family.}
As shown in Table~\ref{tab:imagenet_300}, \OursNet{}-A$0$ is $6\%$ more accurate, has $25\%$ fewer parameters, shows similar runtime, and shows $31\%$ fewer FLOPs than ResNet-$18$ although \OursNet{} has $6$ more layers. Similarly, in contrast to ResNet-$34$, it is more accurate by $3\%$ with $59\%$ fewer parameters, $66\%$ fewer FLOPs, and $23\%$ less layers, while it is fast by $37\%$.
ResNet-$50$ is the widely employed backbone in downstream tasks~\citep{detr, imagenet1hr, maskrcnn, fasterrcnn} due to its affordability regarding representation power, FLOPs, depth, and accuracy. Table~\ref{tab:imagenet_300} shows that \OursNet{}-B$0$ surpasses ResNet-$50$ while being $50\%$ shallower, $22\%$ fewer parameters, $25\%$ fewer FLOPs, and $40\%$ faster.

\noindent\textbf{\OursNet{} \textit{vs} bigger ResNet \citep{resnet} and ResNeXt \citep{resnext} models.} 
As shown in Table~\ref{tab:imagenet_300}, \OursNet{}-C$1$ is better than bigger variants of ResNet, which serves as backbones for cutting-edge works \citep{detr, dndetr}. Our \OursNet{} outperforms them in various aspects while being $72\%$ and $82\%$ less deep relative to ResNet-$101$ and ResNet-$152$, respectively. \OursNet{} also runs faster by $50\%$ in $50\%$ fewer parameters and FLOPs.
In addition, despite being smaller than ResNeXt \citep{resnext}, \OursNet{}-C$1$ outperforms it in various aspects. Overall \OursNet{}-C$1$ is $50\%$ less deeper than ResNeXt-$50$ while running $50\%$ faster at $6\%$ fewer FLOPs, $2\%$ fewer parameters while being more accurate. In contrast to ResNeXt-$101$, \OursNet{}-C$2$ is $75\%$ less deeper, $11\%$ fewer parameters, $12\%$ fewer FLOPs, and $35\%$ faster at a higher accuracy. 

\begin{table}[!t]
\centering


\caption{Comparison with VannilaNet \cite{vanillanet} in training. }
\label{tab:vanillanet_train}
\vspace{-1ex}

\arrayrulecolor{white!60!black}

\scriptsize

{
\setlength{\tabcolsep}{1.5pt}

\begin{tabular}{l c c c c c c c c c}

\midrule

Architecture & \#Depth\bdac{accclr} & \#Epochs\bdac{accclr} & \#Params\bdac{accclr}  & \#FLOPs\bdac{accclr} & Top-$1$ ($\%$)\buac{accclr}  & Train Time Per Epoch \bdac{accclr} & Train Time $300$ Epochs \bdac{accclr} \\


\midrule

\bdota{}~VanillaNet-6 \cite{vanillanet} & $6$ & $300$ & $32.0$M & $6.00$B & $76.36$ & $8$ minutes  & $40$ hours \\

\bdota{}~VanillaNet-8 \cite{vanillanet} & $8$ & $300$ & $37.1$M & $7.70$B  & $79.13$  & $11$ minutes  & $55$ hours \\

\rowcolor{rwclr}
\bdotb{}~\textbf{\OursNet{}-B$\mathbf{1}$} & $26$ & $300$ & $\mathbf{19.8}$M & \textbf{$\mathbf{3.05}$B} & $\mathbf{79.50}$  & $\mathbf{5}$ \textbf{minutes}  & $\mathbf{25}$ \textbf{hours} \\

\midrule

\end{tabular}
}

\vspace{-2ex}
\end{table}

\begin{table}[!t]
\centering

\captionof{table}{\OursNet{} with SE-like modules \cite{senet}.}
\vspace{-1ex}
\label{tab:comnet_with_attention}
\arrayrulecolor{white!60!black}
\scriptsize
\setlength{\tabcolsep}{10.5pt}
\scalebox{\tabscale}
{
\begin{tabular}{l c c c c l}

\toprule
\multicolumn{1}{c}{{\multirow{1}{*}{Approach}}} &  \multicolumn{1}{c}{\#Epochs} &  \multicolumn{1}{c}{\#Depth}  & \multicolumn{1}{c}{{\multirow{1}{*}{\#Params}}}&  \multicolumn{1}{c}{{\#FLOPs}} &  \multicolumn{1}{c}{{Top-$1$ ($\%$)}} \\
\midrule
\bdota{}~ResNet-$50$ + SE \cite{senet}  & $120$ & $50$ & $28.0$M & $4.13$B & $76.85$  \\ 
\bdota{}~ResNet-$50$ + CBAM \cite{cbam} & $120$ & $50$  & $28.0$M & $4.13$B & $77.34$   \\ 
\rowcolor{rwclr}
\bdotb{}~\textbf{\OursNet{}-B$\mathbf{1}$}  & $120$ & $\mathbf{26}$ & $\mathbf{19.2}$M & $\mathbf{3.05}$B & $\mathbf{76.77}$   \\ 
\rowcolor{rwclr}
\bdotb{}~\textbf{\OursNet{}-B$\mathbf{1}$} + SE \cite{senet}  & $120$ & $\mathbf{26}$ & $\mathbf{20.1}$M & $\mathbf{3.10}$B & $\mathbf{77.85}$   \\  \midrule
\bdota{}~ResNet-$50$ + AFF \cite{attentionalfeaturefusion}  & $160$ & $50$ & $30.3$M & $4.30$B & $79.10$  \\ 
\bdota{}~ResNet-$50$ + SKNet \cite{sknet} & $160$  & $50$ & $27.7$M & $4.47$B & $79.21$  \\ 
\rowcolor{rwclr}
\bdotb{}~\textbf{\OursNet{}-C$\mathbf{1}$} + SE \cite{senet} & $160$  & $\mathbf{28}$ & $\mathbf{25.0}$M & $\mathbf{4.13}$B & $\mathbf{79.51}$   \\ 
\bottomrule

\end{tabular}
}

\vspace{-1ex}
\end{table}

\subsection{Additional Experiments}

\noindent\textbf{\OursNet{} has small training walltime.} We provide an additional comparison with the recent ConvNet design, VanillaNet \citep{vanillanet}, under training settings. Table~\ref{tab:vanillanet_train} shows that despite VanillaNet being a shallow network, it has a high training time. We speculate that the large number of channels in the deeper layers of VanillaNet slows down batch processing at large batch sizes. In \OursNet{}, parallel columnar convolutions and controlled parameter growth in the deeper layers counter this issue, leading to lower training time.


\noindent\textbf{\OursNet{} is seamlessly compatible with SE-like \citep{senet} modules.} Table~\ref{tab:comnet_with_attention} shows the results when \OursNet{} is used in conjunction with Squeeze and Excitation (SE) like modules \citep{senet}. It outperforms recent attention mechanism (AFF~\citep{attentionalfeaturefusion}, SKNet~\citep{sknet}, and CBAM~\citep{cbam}) applied to ResNet-50.

\begin{table}[t]
\centering

\caption{\OursNet{} in  state-of-the-art Detection Transformers (DETR) \cite{dndetr} @$12$  epochs setting.}
\label{tab:detection}
\vspace{-1.0ex}
\arrayrulecolor{white!60!black}

\scriptsize

{
\setlength{\tabcolsep}{7pt}

\begin{tabular}{l | c c c | c c c c c c}

\toprule
Method & \#Params  & \#FPS & AP  & AP$_{50}$ & AP$_{75}$ & AP$_{S}$ & AP$_{M}$ & AP$_{L}$
\\

\midrule
\protect\bdota{}~DN-DETR-ResNet$50$ \cite{dndetr} & $44$M  & $24$ &  
$38.3$ & $59.1$ & $41.0$ & $17.3$ & $42.4$ & $57.7$ \\

\rowcolor{rwclr}
\protect\bdotb{}~\textbf{DN-DETR-\OursNet{}-C$\mathbf{2}$}  & $56$M & $\mathbf{25}$ & $39.2$ & $60.0$ & $41.9$ & $18.1$ & $43.0$ & $59.1$ \\

\bottomrule

\end{tabular}
}
\vspace{-1ex}
\end{table}
%

\subsection{\OursNet{} in State-of-the-art Detection Transformer}
We apply \OursNet{} to state-of-the-art object Detection Transformer, DN-DETR \citep{dndetr} to demonstrate the effectiveness of \OursNet{} in the downstream task.
We experiment on MS-COCO \citep{mscoco} benchmark and utilized DN-DETR's default training settings. 

Table~\ref{tab:detection} shows that DN-DETR with \OursNet{} improves the inference speed and average precision compared to the DN-DETR with ResNet-50 backbone. By further optimizing the DETR hyperparameters, \OursNet{} can be configured to deliver better performance.


%
\subsection{Visualization of Attention}
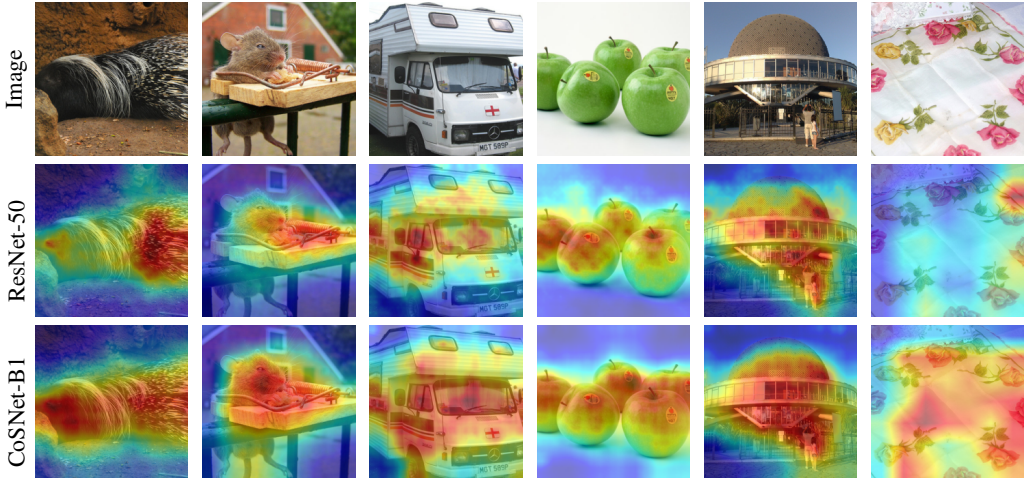
\begin{figure}[!t]

\centering

\FPeval{\imw}{12.8}
\FPeval{\imh}{12.8}

\begin{tikzpicture}

\node (resnet) []{
\tikz{

\node (im1) [xshift=0ex]{\includegraphics[width=\imw ex, height=\imh ex]{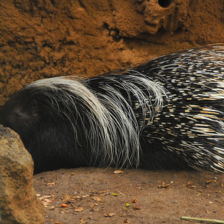}};
\node (im3) [xshift=1*(\imw ex+1.25ex)]{\includegraphics[width=\imw ex, height=\imh ex]{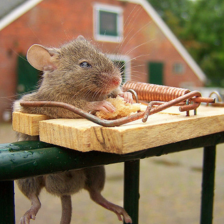}};
\node (im5) [xshift=2*(\imw ex+1.25ex)]{\includegraphics[width=\imw ex, height=\imh ex]{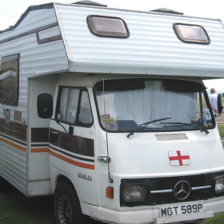}};

\node (im5) [xshift=3*(\imw ex+1.25ex)]{\includegraphics[width=\imw ex, height=\imh ex]{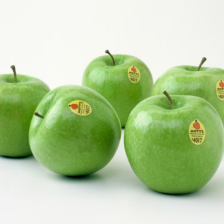}};
\node (im5) [xshift=4*(\imw ex+1.25ex)]{\includegraphics[width=\imw ex, height=\imh ex]{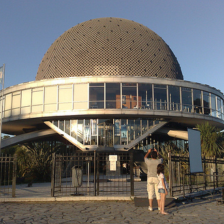}};
\node (im6) [xshift=5*(\imw ex+1.25ex)]{\includegraphics[width=\imw ex, height=\imh ex]{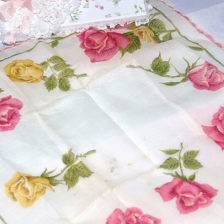}};

\node (im1) [xshift=0ex, yshift=-1*(\imh ex+0.75ex)]{\includegraphics[width=\imw ex, height=\imh ex]{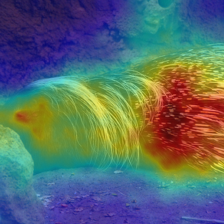}};
\node (im3) [xshift=1*(\imw ex+1.25ex), yshift=-1*(\imh ex+0.75ex)]{\includegraphics[width=\imw ex, height=\imh ex]{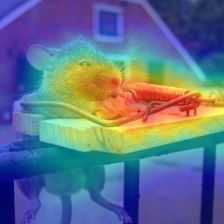}};
\node (im5) [xshift=2*(\imw ex+1.25ex), yshift=-1*(\imh ex+0.75ex)]{\includegraphics[width=\imw ex, height=\imh ex]{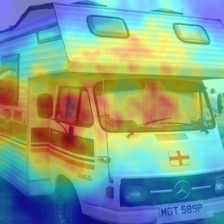}};

\node (im5) [xshift=3*(\imw ex+1.25ex), yshift=-1*(\imh ex+0.75ex)]{\includegraphics[width=\imw ex, height=\imh ex]{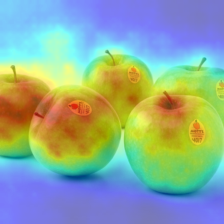}};
\node (im5) [xshift=4*(\imw ex+1.25ex), yshift=-1*(\imh ex+0.75ex)]{\includegraphics[width=\imw ex, height=\imh ex]{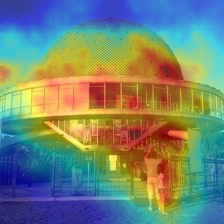}};
\node (im6) [xshift=5*(\imw ex+1.25ex), yshift=-1*(\imh ex+0.75ex)]{\includegraphics[width=\imw ex, height=\imh ex]{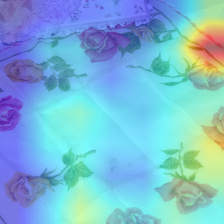}};

\node (im1) [xshift=0ex, yshift=-2*(\imh ex+0.75ex)]{\includegraphics[width=\imw ex, height=\imh ex]{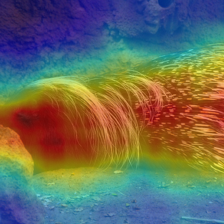}};
\node (im3) [xshift=1*(\imw ex+1.25ex), yshift=-2*(\imh ex+0.75ex)]{\includegraphics[width=\imw ex, height=\imh ex]{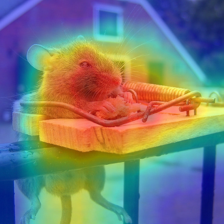}};
\node (im5) [xshift=2*(\imw ex+1.25ex), yshift=-2*(\imh ex+0.75ex)]{\includegraphics[width=\imw ex, height=\imh ex]{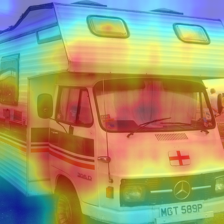}};

\node (im5) [xshift=3*(\imw ex+1.25ex), yshift=-2*(\imh ex+0.75ex)]{\includegraphics[width=\imw ex, height=\imh ex]{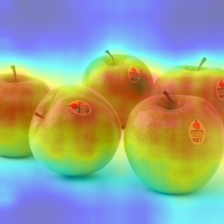}};
\node (im5) [xshift=4*(\imw ex+1.25ex), yshift=-2*(\imh ex+0.75ex)]{\includegraphics[width=\imw ex, height=\imh ex]{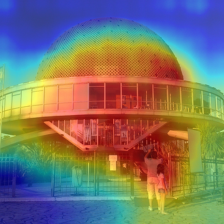}};
\node (im6) [xshift=5*(\imw ex+1.25ex), yshift=-2*(\imh ex+0.75ex)]{\includegraphics[width=\imw ex, height=\imh ex]{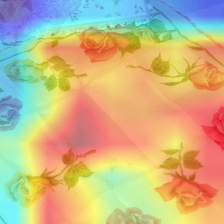}};

\node () [xshift=-8.0ex, yshift=0.0ex, rotate=90, scale = 0.9]{Image};
\node () [xshift=-8.0ex, yshift=-14.5ex, rotate=90, scale = 0.9]{ResNet-$50$};
\node () [xshift=-8.0ex, yshift=-28.0ex, rotate=90, scale = 0.9]{\OursNet{}-B$1$};
}
};

\end{tikzpicture}
\vspace{-4.0ex}
\caption{CAM \cite{fullgradcam} visualizations. Notably, \OursNet{} attends the class regions more accurately than the baseline.}
\label{fig:gradcam}

\vspace{-2ex}
\end{figure}
%

%
To comprehend \OursNet{}'s better performance, we investigate its class activation maps (CAM) on ImageNet \citep{imagenet} validation set. We use CAM output from popular Full-Grad-CAM \citep{fullgradcam} for a given class. CAM visualizations of ResNet-$50$ and \OursNet{}-B$1$ are shown in Figure~\ref{fig:gradcam}. It can be seen that \OursNet{}, despite being $50\%$ shallower than ResNet, is better at learning to attend regions of the target class relative to the baseline. 


\section{Conclusion}
\label{sec:conc}
%

We propose \textit{\OursNet{}}, which revisits ConvNet design based on multiple aspects for concise models. \OursNet{} is based on our parallel columnar convolutions and input replication concepts to be efficient in parameters, FLOPs, accuracy, latency, and training duration. Through extensive experimentation and ablations, we show that \OursNet{} rivals many representative ConvNets and ViTs such as ResNet, ResNeXt, RegNet, RepVGG, and ParNet, VanillaNet, DeiT, EfficientViT while being shallower, faster, and being architecturally simpler.

\noindent\textbf{Future work.} \OursNet{} is open for improvement. 
In this paper, we have built a simple template architecture that can further evolve like ConvNext~\citep{convnext}. 
For instance, a comprehensive design space of \OursNet{} including mobile regime can be explored, similar to RegNet~\citep{regnet}. Besides, layer merging post-training, shown in VanillarNet~\citep{vanillanet}, can be utilized to develop shallower variants of \OursNet{}. In addition to that, \OursNet{} can also be married with a Transformer attention mechanism like \citep{coatnet} or \citep{efficientvit}.

\noindent\textbf{Acknowledgements.} Jaesik Park was supported by MSIT grant (RS-2021-II211343: AI Graduate School Program at Seoul National University (5\%) and 2023R1A1C200781211 (95\%))



\bibliographystyle{iclr2025_conference}
\bibliography{bibfile}

\newpage

\appendix
\section*{Appendix}
\setcounter{figure}{0}
\setcounter{table}{0}
\renewcommand{\thefigure}{A\arabic{figure}}
\renewcommand{\thetable}{A\arabic{table}}
\section{\OursNet{} Instances}
Table~\ref{tab:designspace} shows \OursNet{} instances configurations mentioned in the main paper.

\begin{table}[!h]
\centering

\caption{\OursNet{} instances Configurations.}
\label{tab:designspace}

\arrayrulecolor{white!60!black}

\scriptsize

\setlength{\tabcolsep}{4.0pt}

\scalebox{\tabscale}{
\begin{tabular}{c c c c c c c c c c c c c c c c c c c c c c c}

\toprule

Model & \multicolumn{4}{c}{{$P_c$}} & \multicolumn{4}{c}{{$N$}}  & \multicolumn{4}{c}{{$l$}}  & \multicolumn{4}{c}{{$M$}}  &  \multicolumn{1}{c}{{\#Depth}}  & \multicolumn{1}{c}{{\#Params}} & \multicolumn{1}{c}{{\#FLOPs}} \\

\midrule

\multicolumn{1}{c|}{\bdotb{}~\OursNet-A$0$} & $256$ & $512$ & $1024$ &$2048$ & \multicolumn{1}{|c}{$16$} & $32$ & $64$ & $128$  &  \multicolumn{1}{|c}{$3$} & $4$ & $6$ & $3$  &  \multicolumn{1}{|c}{$1$} & $1$ & $1$ & $1$  & \multicolumn{1}{|c}{$26$} &  \multicolumn{1}{c}{$8.8$M} & $1.25$B  \\

\multicolumn{1}{c|}{\bdotb{}~\OursNet-A$1$} & $256$ & $512$ & $1024$ &$2048$ & \multicolumn{1}{|c}{$16$} & $32$ & $64$ & $128$  & \multicolumn{1}{|c}{$3$} & $4$ & $6$ & $3$  &  \multicolumn{1}{|c}{$4$} & $4$ & $4$ & $4$ & \multicolumn{1}{|c}{$26$} &  $12.1$M   & $1.77$B \\

\midrule

\multicolumn{1}{c|}{\bdotb{}~\OursNet-B$0$} & $256$ & $512$ & $1024$ &$2048$ & \multicolumn{1}{|c}{$32$} & $64$ & $128$ & $256$  &  \multicolumn{1}{|c}{$3$} & $4$ & $6$ & $3$  &  \multicolumn{1}{|c}{$4$} & $4$ & $4$ & $4$ & \multicolumn{1}{|c}{$26$} &  $19.8$M  & $3.05$B  &  \\ 

\multicolumn{1}{c|}{\bdotb{}~\OursNet-B$1$} & $256$ & $512$ & $1024$ &$2048$ & \multicolumn{1}{|c}{$32$} & $64$ & $128$ & $256$  &  \multicolumn{1}{|c}{$3$} & $4$ & $6$ & $3$  &  \multicolumn{1}{|c}{$5$} & $5$ & $5$ & $5$ & \multicolumn{1}{|c}{$26$} &  $22.6$M  & $3.51$B  \\

\multicolumn{1}{c|}{\bdotb{}~\OursNet-B$2$} & $256$ & $512$ & $1024$ &$2048$ & \multicolumn{1}{|c}{$32$} & $64$ & $128$ & $256$  &  \multicolumn{1}{|c}{$3$} & $4$ & $6$ & $3$  &  \multicolumn{1}{|c}{$4$} & $4$ & $16$ & $4$ & \multicolumn{1}{|c}{$26$} &  $30.0$M  & $5.1$B &  \\ 

\midrule

\multicolumn{1}{c|}{\bdotb{}~\OursNet-C$1$} & $256$ & $512$ & $1024$ &$2048$ & \multicolumn{1}{|c}{$48$} & $80$ & $144$ & $272$  &  \multicolumn{1}{|c}{$4$} & $4$ & $6$ & $4$  &  \multicolumn{1}{|c}{$4$} & $4$ & $4$ & $4$ & \multicolumn{1}{|c}{$28$} & $24.4$M  & $4.12$B  &  \\

\multicolumn{1}{c|}{\bdotb{}~\OursNet-C$2$} & $256$ & $512$ & $1024$ &$2048$ & \multicolumn{1}{|c}{$48$} & $80$ & $144$ & $272$  &  \multicolumn{1}{|c}{$3$} & $4$ & $6$ & $3$  &  \multicolumn{1}{|c}{$6$} & $6$ & $16$ & $6$ & \multicolumn{1}{|c}{$26$} &  $38.9$M  & $7.09$B &  \\

\bottomrule

\end{tabular}
}
\vspace{-3.0ex}
\end{table}
\begin{table*}[t]
\centering

\caption{Effect of parallel columnar convolution (PCC), \# of kernels $N$, \# of layers $l$, \# of parallel convolutions $M$, and Deeper projections (DP). Values of $M, N, l$ are for each of the four \OursNet{} stages. Ablations are conducted at $120$ epochs.}
\label{tab:effacm}
%
%

\arrayrulecolor{white!60!black}

\tiny

\setlength{\tabcolsep}{3.5pt}

\scalebox{1.10}{
\begin{tabular}{c c c c c c c c c c c c c c c c c c c}

\toprule

Row & \multicolumn{4}{c}{{$N$}} & \multicolumn{4}{c}{{$l$}}  & \multicolumn{4}{c}{{$M$}} &  \multicolumn{1}{c}{{\#Depth}}  & \multicolumn{1}{c}{\texttt{DP}} &  \multicolumn{1}{c}{Residual in \texttt{PCC}} & \multicolumn{1}{c}{{\#Params}} & \multicolumn{1}{c}{{\#FLOPs}}   & \multicolumn{1}{c}{Top-$1$ ($\%$)} \\
\midrule

\multicolumn{1}{c|}{R$0$} & \bdotb{}~$16$ & $32$ & $64$ & $128$ &  \multicolumn{1}{|c}{$3$} & $4$ & $6$ & $3$  &  \multicolumn{1}{|c}{$1$} & $1$ & $1$ & $1$  & \multicolumn{1}{|c}{$26$} &  \multicolumn{1}{c}{\cmark}  & \cmark  & \multicolumn{1}{c}{$8.80$M} & $1.25$B & \multicolumn{1}{c}{$74.45$} \\
\multicolumn{1}{c|}{R$1$} & \bdotb{}~$16$ & $32$ & $64$ & $128$ &  \multicolumn{1}{|c}{$3$} & $4$ & $6$ & $3$  &  \multicolumn{1}{|c}{$4$} & $4$ & $4$ & $4$  & \multicolumn{1}{|c}{$26$} &  \multicolumn{1}{c}{\cmark}  & \cmark  & \multicolumn{1}{c}{$12.1$M} & $1.77$B & \multicolumn{1}{c}{$75.65$} \\
\multicolumn{1}{c|}{R$2$} & \bdotb{}~$16$ & $32$ & $64$ & $128$ &  \multicolumn{1}{|c}{$3$} & $4$ & $6$ & $3$  &  \multicolumn{1}{|c}{$5$} & $5$ & $5$ & $5$  & \multicolumn{1}{|c}{$26$} &  \multicolumn{1}{c}{\cmark}  & \cmark  & \multicolumn{1}{c}{$13.2$M} & $1.95$B & \multicolumn{1}{c}{$76.01$} \\ \midrule
\multicolumn{1}{c|}{R$3$} & \bdotb{}~$32$ & $64$ & $128$ & $256$ & \multicolumn{1}{|c}{$3$} & $4$ & $6$ & $3$  &  \multicolumn{1}{|c}{$1$} & $1$ & $1$ & $1$ & \multicolumn{1}{|c}{$26$} &  \multicolumn{1}{c}{\cmark}  & \cmark & \multicolumn{1}{c}{$11.3$M} & $1.65$B & \multicolumn{1}{c}{$75.37$}\\ 
\multicolumn{1}{c|}{R$4$} & \bdotb{}~$32$ & $64$ & $128$ & $256$  &  \multicolumn{1}{|c}{$3$} & $4$ & $6$ & $3$  &  \multicolumn{1}{|c}{$4$} & $4$ & $4$ & $4$   & \multicolumn{1}{|c}{$26$} &  \multicolumn{1}{c}{\cmark}  & \cmark  & \multicolumn{1}{c}{$19.8$M} & $3.05$B & \multicolumn{1}{c}{$76.76$} \\
\multicolumn{1}{c|}{R$5$} & \bdotb{}~$32$ & $64$ & $128$ & $256$  &  \multicolumn{1}{|c}{$3$} & $4$ & $6$ & $3$  &  \multicolumn{1}{|c}{$5$} & $5$ & $5$ & $5$  & \multicolumn{1}{|c}{$26$} &  \multicolumn{1}{c}{\cmark}  & \cmark  & \multicolumn{1}{c}{$22.6$M} & $3.51$B & \multicolumn{1}{c}{$77.01$} \\
%
\midrule
\multicolumn{1}{c|}{R$6$} & \bdotb{}~$32$ & $64$ & $128$ & $256$  &  \multicolumn{1}{|c}{$3$} & $4$ & $6$ & $3$  &  \multicolumn{1}{|c}{$1$} & $1$ & $1$ & $1$  & \multicolumn{1}{|c}{$26$} &  \multicolumn{1}{c}{\cmark}  & \xmark & \multicolumn{1}{c}{$11.3$M} & $1.65$B & \multicolumn{1}{c}{$75.28$}\\ 
\multicolumn{1}{c|}{R$7$} & \bdotb{}~$32$ & $64$ & $128$ & $256$  & \multicolumn{1}{|c}{$3$} & $4$ & $6$ & $3$  &  \multicolumn{1}{|c}{$1$} & $1$ & $1$ & $1$ &   \multicolumn{1}{|c}{$26$} &  \multicolumn{1}{c}{\cmark}  & \cmark & \multicolumn{1}{c}{$11.3$M} & $1.65$B & \multicolumn{1}{c}{$75.37$}\\   

\midrule
\multicolumn{1}{c|}{R$8$} & \bdotb{}~$32$ & $64$ & $128$ & $256$  & \multicolumn{1}{|c}{$4$} & $5$ & $20$ & $3$  &  \multicolumn{1}{|c}{$1$} & $1$ & $1$ & $1$ &  \multicolumn{1}{|c}{$44$} &  \multicolumn{1}{c}{\cmark}  & \xmark & \multicolumn{1}{c}{$13.4$M} & $2.12$B & \multicolumn{1}{c}{$75.18$} \\
\multicolumn{1}{c|}{R$9$} & \bdotb{}~$32$ & $64$ & $128$ & $256$ & \multicolumn{1}{|c}{$4$} & $5$ & $20$ & $3$  &  \multicolumn{1}{|c}{$1$} & $1$ & $1$ & $1$ & \multicolumn{1}{|c}{$44$} &  \multicolumn{1}{c}{\cmark}  & \cmark &  \multicolumn{1}{c}{$13.4$M} & $2.12$B & \multicolumn{1}{c}{$75.88$} \\  

\midrule

\multicolumn{1}{c|}{R$10$} & \bdotb{}~$32$ & $64$ & $128$ & $256$  &  \multicolumn{1}{|c}{$3$} & $4$ & $6$ & $3$  &  \multicolumn{1}{|c}{$1$} & $1$ & $1$ & $1$   & \multicolumn{1}{|c}{$26$} &  \multicolumn{1}{c}{\xmark}  & \cmark &  \multicolumn{1}{c}{$8.5$M} & $1.29$B & \multicolumn{1}{c}{$73.61$} \\
\multicolumn{1}{c|}{R$11$} & \bdotb{}~$32$ & $64$ & $128$ & $256$ & \multicolumn{1}{|c}{$3$} & $4$ & $6$ & $3$  &  \multicolumn{1}{|c}{$1$} & $1$ & $1$ & $1$  & \multicolumn{1}{|c}{$26$} &  \multicolumn{1}{c}{w/o. \texttt{Pooling}}  & \cmark &  \multicolumn{1}{c}{$9.8$M} & $1.44$B & \multicolumn{1}{c}{$74.15$}\\  
\multicolumn{1}{c|}{R$12$} & \bdotb{}~$32$ & $64$ & $128$ & $256$ & \multicolumn{1}{|c}{$3$} & $4$ & $6$ & $3$  &  \multicolumn{1}{|c}{$1$} & $1$ & $1$ & $1$  & \multicolumn{1}{|c}{$26$} &  \multicolumn{1}{c}{w. \texttt{Pooling}}  & \cmark & \multicolumn{1}{c}{$9.8$M} & $1.44$B & \multicolumn{1}{c}{$75.37$}\\

 \midrule
\multicolumn{1}{c|}{R$13$} & \bdotb{}~$32$ & $64$ & $128$ & $256$  &  \multicolumn{1}{|c}{$3$} & $4$ & $6$ & $3$  &  \multicolumn{1}{|c}{$4$} & $4$ & $4$ & $4$   & \multicolumn{1}{|c}{$26$} &  \multicolumn{1}{c}{\cmark}  & \cmark  & \multicolumn{1}{c}{$19.8$M} & $3.51$B & \multicolumn{1}{c}{$76.76$} \\
\multicolumn{1}{c|}{R$14$} & \bdotb{}~$32$ & $32$ & $32$ & $32$  &  \multicolumn{1}{|c}{$3$} & $4$ & $6$ & $3$  &  \multicolumn{1}{|c}{$4$} & $8$ & $16$ & $32$  & \multicolumn{1}{|c}{$26$} &  \multicolumn{1}{c}{\cmark}  & \cmark  & \multicolumn{1}{c}{$18.4$M} & $3.42$B & \multicolumn{1}{c}{$71.20$} \\

\bottomrule

\end{tabular}
}
%
\end{table*}
\section{Ablation Study}
\label{sec:ablations}
%
%
\par
\textbf{Varying $\mathbf{M}$ and $\mathbf{N}$.}
Table~\ref{tab:effacm} demonstrates the effect of varying $N$ and $M$ (R$0$-R$5$). We first fix the values of $N$ and vary $M$ (R$0$-R$5$), and then vary $M$ while fixing $N$ (R$0$~$\leftrightarrow$~R$3$, R$1$~$\leftrightarrow$~R$4$, R$2$~$\leftrightarrow$~R$5$). For fixed $N$, accuracy improves by increasing $M$, and the same effect is seen by fixing $M$ while varying $N$. 
It can be noticed that parameters, FLOPs can be controlled by changing the $M$ (R$1$~$\leftrightarrow$~R$2$, R$4$~$\leftrightarrow$~R$5$), which directly reflects accuracy.
\par
\textbf{Effect of \texttt{PCC}.}
We compare instances having different $N, M$, but have similar parameters and FLOPs budget, for instance, R$1$~$\leftrightarrow$~R$2$, R$1$~$\leftrightarrow$~R$3$, Table~\ref{tab:effacm}. Noticeably, $R2$ with $5$ \texttt{PCC} is better by $0.36\%$ in accuracy, only at $1.1$M more parameters relative to R$1$. Similarly, $R1$ is better by $0.28\%$ in accuracy, only at $0.8$M more parameters relative to R$3$. It shows that multiple \texttt{PCC}s facilitates improved accuracy in just a fraction of parameters and FLOPs. Moreover, if comparing R$9$ (a deeper model) with R$2$, R$2$ achieves $0.13\%$ more accuracy in $0.2$M fewer parameters and $0.17$B fewer FLOPs. It shows the advantage of \emph{having multiple convolutional modules while being shallower}.
\par
\textbf{Varying $\mathbf{l}$.}
The impact of varying $l$ is shown in R$9$, Table~\ref{tab:effacm}. It can be seen that \emph{going deeper is not necessary} because a shallower version with same parameters (R$2$) is more accurate. Moreover, increased depth causes increased latency in R$9$. Therefore, we stick to $20-40$ layers of depth.

\textbf{Group Convolution or ResNext-like \cite{resnext} Setting}
We also conduct additional experiments where each \texttt{PCC} is followed by a $1 \times1$ convolution as done in group convolutions while keeping depth and parameters constant. We observe a $1\%$ accuracy drop. This indicates that frequent fusion similar to ResNeXt is not necessary.
\par
\textbf{Effect of Shallow Projections in \textbf{\texttt{PCC}}.}
R$6$-R$9$, Table~\ref{tab:effacm} shows this analysis. For the shallower model, the residual connection shows only minor improvement ($0.09\%$), however, for the deeper model, the effect of residual connections is noticeable ($0.70\%$). 
\par
\textbf{Effect of Deep Projections (\texttt{DP}).}
We train an \OursNet{} instance in three ways: \textit{First}, remove \texttt{DP} entirely, \textit{Second}, use \texttt{DP} without \texttt{pooling}, and \textit{Third}, \texttt{DP} with \texttt{pooling}. See Table~\ref{tab:effacm} for the analysis. It can be noticed that without \texttt{DP} (R$10$), the model suffers with heavy accuracy loss of $\sim 0.54\%$ relative to when \texttt{DP} is used without \texttt{pooling} (R$11$). Moreover, when using \texttt{DP} with \texttt{pooling} (R$12$), accuracy improves, i.e., $1.22\%$ and $1.76\%$ relative to R$11$ and R$10$, respectively, because \texttt{pooling} provides more spatial context to the $1 \times1$ $L_p$ layer by summarizing the neighborhood.

\par
\textbf{Effect of using very small $N$ to compare with Group Convs \cite{shufflenetv1} and depthwise Conv-like \cite{mobilenetv1} structure.}
From Table~\ref{tab:effacm}, R$13$-$14$, it can be seen that when in deeper layers, $N$ is restricted to a very smaller value while keeping parameters or FLOPs the same, accuracy decreases considerably. This is because of the reason mentioned in the main paper (Sec.``Fuse-Once'') that too few connections restrict the learning ability of neurons. Hence, they need frequent fusion similar to GroupWise and Depthwise convolution methods, but it increases depth. To avoid that, we increase $N$ as we go deep down in \OursNet{}, which does not require frequent fusion due to a sufficiently large number of neuron connections. Thus, we fuse only once, eliminating the need for fusion $1 \times 1$ layers, thus smaller depth and lower latency.

\section{The effect of batch sizes of the baseline approaches.}

In the literature, some baselines are trained with larger batch sizes (above 1024), but others have been trained at a much smaller batch size ($256$). 
Therefore, we retrained high batch size baselines with $256$ batch sizes to avoid getting biased conclusions about the effects of large batch sizes. Such results with 256 batch size are carefully reported in Table~\ref{tab:imagenet_300}.

In this section, we present the results of the baselines with larger batch sizes in Table~\ref{tab:eff_batch_size}. As widely studied, the baseline approaches~\cite{efficientnet,efficientvit,convnext} show improved accuracy. 
Interestingly, it can be noticed that \OursNet{} trained with a 256 batch size can compete with state-of-the-art approaches trained with a larger batch size. This shows the utility of obtaining higher accuracies in resource-constrained training scenarios (i.e., limited memory to fit 4096 batch, etc.).

\begin{table*}[!t]
\centering

\caption{Effect of batch size on the baselines and \OursNet{} in the context.}
\label{tab:eff_batch_size}
\vspace{-1ex}
\arrayrulecolor{white!60!black}
\setlength{\tabcolsep}{4.0pt}

\scriptsize

\resizebox{0.995\linewidth}{!}
{

\begin{tabular}{l | c | c r r r r c c}

\toprule

Architecture & Type & Batch Size & \#Depth~\bdac{accclr} & \#Params~\bdac{accclr} & FLOPs~\bdac{accclr} & Latency~\bdac{accclr} & FPS~\buac{accclr} & Top-$1$ ($\%$)~\buac{accclr} \\

\midrule

\bdota{}~EfficientNet-B$0$ \cite{efficientnet} &  ConvNet & $256$ & $49$ &  $5.3$M & $0.40$B & $8$ms & $125$ & $75.1$ \\

\bdota{}~EfficientNet-B$0$ \cite{efficientnet} &  ConvNet & $2048$ & $49$ &  $5.3$M & $0.40$B & $8$ms & $125$ & $77.1$ \\

\bdota{}~EfficientViT-M$5$ \cite{efficientvit} & Transformer & $256$ & $70$ &  $12.4$M & $0.60$B & $7$ms & $142$ & $76.8$ \\

\bdota{}~EfficientViT-M$5$ \cite{efficientvit} & Transformer & $2048$ & $70$ &  $12.4$M & $0.60$B & $7$ms & $142$ & $77.1$ \\

\bdotb{}~\textbf{\OursNet{}-A$0$} & ConvNet & $256$ & $26$ &  $8.8$M & $1.25$B & $6$ms & $167$ & $77.1$ \\

\bdotb{}~\textbf{\OursNet{}-A$1$-PFF} & ConvNet & $256$ & $38$ & $12.7$M & $1.93$B  & $7$ms & $143$ & $79.7$ \\ 

\midrule

\bdota{}~ConvNeXt-T \cite{convnext} & ConvNet & $256$ &  $59$ & $29.0$M & $4.50$B & $13$ms & $77$ & $81.8$ \\

\bdota{}~ConvNeXt-T \cite{convnext} & ConvNet & $4096$ &  $59$ & $29.0$M & $4.50$B & $13$ms & $77$ & $82.1$ \\

\bdotb{}~\textbf{\OursNet{}-C$2$} & ConvNet & $256$  & $26$ & $38.9$M & $7.09$B & $11$ms & $90$ & $82.1$ \\

\bdotb{}~\textbf{\OursNet{}-B$2$-PFF} & ConvNet & $256$  & $38$ & $34.3$M & $5.91$B  & $10$ms & $100$ & $82.7$ \\


\bottomrule

\end{tabular}
}
%
\end{table*}

\begin{table}[t]
\centering

\caption{\OursNet{} in  RetinaNet $x1$ \cite{retinanet} object detection framework.}
\label{tab:additional_detection}
\vspace{-1.0ex}
\arrayrulecolor{white!60!black}

\scriptsize

{
\setlength{\tabcolsep}{7pt}

\begin{tabular}{l | c c c | c c c c c c}

\toprule
Method & \#Depth  & \#Params & AP  & AP$_{S}$ & AP$_{M}$ & AP$_{L}$
\\
\midrule
\protect\bdota{}~EfficientViT-M$4$ \cite{efficientvit} & $42$  & $8.8$M &  
$32.7$ & $17.6$ & $35.3$ & $46.0$ \\

\rowcolor{rwclr}
\protect\bdotb{}~\textbf{\OursNet{}-A$\mathbf{0}$}  & $\mathbf{26}$ & $8.8$M & $34.3$ & $19.1$ & $38.0$ & $49.1$ \\

\bottomrule

\end{tabular}
}
\vspace{-1ex}
\end{table}
%

\begin{table}[t]
\centering

\caption{\OursNet{} in  PSPNet \cite{pspnet} semantic segmentation framework.}
\label{tab:segmentation}
\vspace{-1.0ex}
\arrayrulecolor{white!60!black}

\scriptsize

{
\setlength{\tabcolsep}{7pt}

\begin{tabular}{l | c c c}

\toprule
Method & \#Params & mIoU  & FPS
\\
\midrule 
\protect\bdota{}~RepVGG-B$1$g$2$ \cite{repvgg}	& $41.36$M & $78.88$ & $13$ \\
\protect\bdota{}~ResNet-$50$ &	$25.5$M &	$77.17$ &	$13$ \\

\rowcolor{rwclr}
\protect\bdotb{}~\textbf{CosNet-B1} & $22.0$M & $79.05$ & $17$ \\

\bottomrule

\end{tabular}
}
\vspace{-1ex}
\end{table}
%

\section{Additional Results}

Table~\ref{tab:additional_detection} shows results on RetinaNet $x1$ \cite{retinanet} detection pipeline. It can be seen that, for a comparable vision transformer backbone, \OursNet{} performs better.
We also provide semantic segmentation results for the popular PSPNet {pspnet} semantic segmentation framework. It can be seen that \OursNet{} performs better than the baselines.

\section{Training setting}
\label{sec:trainingsetting}
We train models in PyTorch \cite{pytorch} using eight NVIDIA A40 GPUs.

%
%
%

\section{Complete Network Visualization}
\begin{figure}[htb]
\centering
\begin{tikzpicture}

\node (unit)[]{
\includegraphics[scale=0.4]{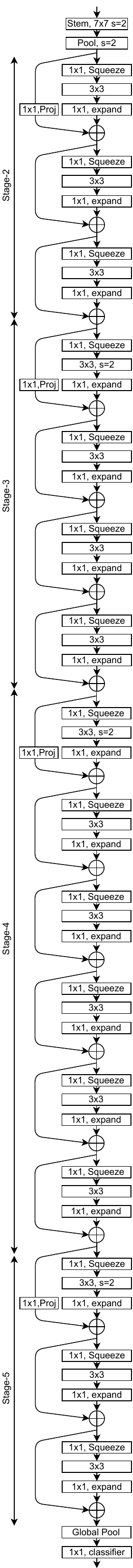}
};
\node (unit1)[xshift=30ex]{
\includegraphics[scale=0.4]{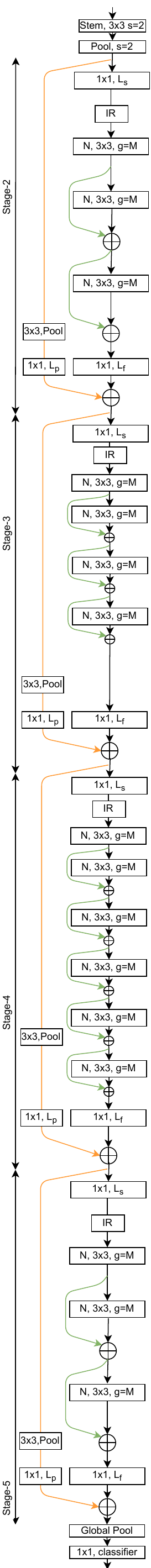}
};
\node (a) [below of=unit,xshift=-12ex, yshift=-1ex]{\scriptsize (a) ResNet-50};
\node (b) [below of=unit1,xshift=-12ex, yshift=-1ex]{\scriptsize (b) \OursNet{}-B1};

\end{tikzpicture}
\subfloat{\label{fig:resnet50}}
\subfloat{\label{fig:cosnetb1}}
\vspace{-0.0ex}
\caption{Illustration of (a) ResNet-50 \cite{resnet} network, and (b) \OursNet{}-B1. It must be noted that by merely replacing the residual bottleneck-based stages of ResNet with the proposed \OursNet{}-unit, our \OursNet{} variant becomes roughly $50\%$ less deep, has $22\%$ fewer parameters, $25\%$ fewer FLOPs, and runs $40\%$ faster. It shows the utility of \OursNet{} design from an efficiency perspective in multiple aspects.}
\label{fig:complete_network}
\vspace{-3.5ex}
\end{figure}

We also visualize the complete architecture of \OursNet{}-B1 variant and have put it in the context of ResNet-like models. We have plotted ResNet-$50$ variant. Please see Figure~\ref{fig:complete_network}.

\end{document}